\documentclass[journal]{IEEEtran}
\usepackage{multicol} 
\usepackage{amsmath,amsfonts}
\usepackage{algorithmic}
\usepackage{algorithm}
\usepackage{array}
\usepackage[caption=false,font=normalsize,labelfont=sf,textfont=sf]{subfig}
\usepackage{textcomp}
\usepackage{stfloats}
\usepackage{url}
\usepackage[colorlinks,linkcolor=blue,citecolor=blue]{hyperref}
\usepackage{verbatim}
\usepackage{graphicx} 
\usepackage{booktabs}
\usepackage{multirow}    
\usepackage{amsmath}     
\usepackage{cite}
\usepackage[inkscapelatex=false]{svg}
\usepackage{bm}
\begin{document}

\title{UGAE: Unified Geometry and Attribute Enhancement for G-PCC Compressed Point Clouds}

\author{Pan Zhao, Hui Yuan,~\IEEEmembership{Senior Member,~IEEE,} Chongzhen Tian, Tian Guo, Raouf Hamzaoui,~\IEEEmembership{Senior Member,~IEEE,} and Zhigeng Pan
\thanks{This work was supported in part by the National Natural Science Foundation
of China under Grants 62222110, 62172259, and 62072150, the Taishan Scholar Project of
Shandong Province (tsqn202103001), the Shandong Provincial Natural Science Foundation under Grant ZR2022ZD38, and the OPPO Research Fund. (Corresponding author: Hui Yuan)}
\thanks{Pan Zhao, Hui Yuan, ChongZhen Tian, and Tian Guo are with the School of Control Science and
Engineering, Shandong University, Jinan 250061, China, and also with
the Key Laboratory of Machine Intelligence and System Control, Ministry
of Education, Ji’nan, 250061, China (e-mail: panz@mail.sdu.edu.cn;
huiyuan@sdu.edu.cn; 202420789@mail.sdu.edu.cn; guotiansdu@mail.sdu.edu.cn)

Raouf Hamzaoui is with the School of Engineering and Sustainable
Development, De Montfort University, LE1 9BH Leicester, UK. (e-mail:
rhamzaoui@dmu.ac.uk)

Zhigeng Pan is with the Institute of NUIST-MetaX for Graphics Processing Unit, Nanjing University of Information Science and Technology Nanjing, China (e-mail: 443922077@qq.com)
}}

\markboth{Journal of \LaTeX\ Class Files,~Vol.~14, No.~8, june~2025}%
{Zhao\MakeLowercase{\textit{et al.}}: UGAE: Unified Geometry and Attribute Enhancement for G-PCC Compressed Point Clouds}


\maketitle

\begin{abstract}
Lossy compression of point clouds reduces storage and transmission costs; however, it inevitably leads to irreversible distortion in geometry structure and attribute information. To address these issues, we propose a unified geometry and attribute enhancement (UGAE) framework, which consists of three core components: post-geometry enhancement (PoGE), pre-attribute enhancement (PAE), and post-attribute enhancement (PoAE). In PoGE, a Transformer-based sparse convolutional U-Net is used to reconstruct the geometry structure with high precision by predicting voxel occupancy probabilities. Building on the refined geometry structure, PAE introduces an innovative enhanced geometry-guided recoloring strategy, which uses a detail-aware K-Nearest Neighbors (DA-KNN) method to achieve accurate recoloring and effectively preserve high-frequency details before attribute compression. Finally, at the decoder side, PoAE uses an attribute residual prediction network with a weighted mean squared error (W-MSE) loss to enhance the quality of high-frequency regions while maintaining the fidelity of low-frequency regions. UGAE significantly outperformed existing methods on three benchmark datasets: 8iVFB, Owlii, and MVUB. Compared to the latest G-PCC test model (TMC13v29), UGAE achieved an average BD-PSNR gain of 9.98 dB and \textbf{90.98\%} BD-bitrate savings for geometry under the D1 metric, as well as a 3.67 dB BD-PSNR improvement with \textbf{56.88\%} BD-bitrate savings for attributes on the Y component. Additionally, it improved perceptual quality significantly. Our source code will be released on GitHub at: \url{https://github.com/yuanhui0325/UGAE}

\end{abstract}


\section{Introduction}
\IEEEPARstart{P}{oint} clouds, as sets of unstructured points in 3D space, can accurately represent the detailed features of object surfaces through their geometry coordinates and multidimensional attributes (such as color, reflectivity, etc.). With the advancement of 3D acquisition technologies, current methods for obtaining point clouds encompass LiDAR scanning and multi-view reconstruction~\cite{ref1}, significantly expanding their potential and value in cutting-edge applications such as autonomous driving~\cite{ref2}, cultural heritage preservation~\cite{ref3}, and virtual reality~\cite{ref4}. However, high-precision point clouds typically consist of a large number of sampling points, which presents significant challenges for data storage and transmission.

To efficiently compress point clouds, the Moving Picture Experts Group (MPEG) has developed two standards: Geometry-based Point Cloud Compression (G-PCC)~\cite{ref5} and Video-based Point Cloud Compression (V-PCC)~\cite{ref6}. G-PCC compresses 3D point clouds by directly encoding their geometry and attributes, whereas V-PCC projects them onto 2D images and uses conventional 2D video coders for compression. In recent years, with the rapid development of deep learning, learning-based methods~\cite{ref7, ref8, ref9, ref10, ref11, ref12, ref13, ref14, ref15, ref16, ref17, ref18, ref19, ref20, ref21, ref22, ref23, ref24, ref25, ref26, ref27, ref28, ref29, ref30, ref31, ref32, ref33, ref34, ref35, ref36} have demonstrated superior rate-distortion (R-D) performance for both geometry and attribute compression of point clouds. Although existing compression methods have significantly reduced data volume, they inevitably introduce distortion in geometry and attributes during the compression process. Therefore, balancing compression efficiency with visual fidelity remains a challenging issue.

To mitigate distortion, methods such as PU-Net~\cite{ref37}, PUFA-GAN~\cite{ref38}, and PU-Mask~\cite{ref39} recover geometry by increasing the number of points through multi-level feature extraction; however they are primarily designed for small-scale point clouds. For large-scale scenarios, methods such as PU-Dense~\cite{ref40}, GRNet~\cite{ref41}, and G-PCC++~\cite{ref42} use sparse convolutions~\cite{ref43}, which significantly improve the processing capability for large point clouds. However, these methods focus exclusively on geometry enhancement and overlook the recovery of attribute information. Most existing methods~\cite{ref44, ref45, ref46} assume lossless geometry compression and estimate attribute residuals using the reconstructed point cloud. Although G-PCC++~\cite{ref42} attempts to jointly consider both geometry and attribute distortions to some extent, its joint strategy is primarily embodied in the attribute enhancement stage: using the enhanced geometry to assist the interpolation of G-PCC reconstruction attributes for subsequent enhancement—a process similar to super-resolution tasks in images. However, this joint strategy is only applied during the reconstruction phase and lacks collaborative optimization throughout the entire compression procedure, meaning that the benefits of geometry enhancement on attribute compression and reconstruction are not fully exploited.

\IEEEpubidadjcol
Most existing point cloud compression methods sequentially compress geometry and attributes, but this approach suffers from two critical issues: (1) geometry quantization causes structural distortion, and (2) attribute compression based on the lossy geometry leads to error accumulation, resulting in irreversible loss, especially for high-frequency details. To address these problems, we propose a unified geometry and attribute enhancement (UGAE) framework that consists of three components: post-geometry enhancement (PoGE), pre-attribute enhancement (PAE), and post-attribute enhancement (PoAE). In PoGE, we propose a U-Net~\cite{ref47} architecture with Transformer blocks~\cite{ref48}, which takes the lossy geometry as input to extract multi-scale geometry features. These features are then fed into the geometry enhancement head, where they are upsampled using a transpose sparse convolution (TSConv) layer and enhanced through dense connections~\cite{ref49}. The network outputs the Top-K high-probability voxels to improve the quality of the lossy geometry. Additionally, to ensure output consistency and eliminate randomness caused by GPU parallelism, we move the execution of the TSConv layer operations to the CPU. Next, we introduce PAE, which uses the enhanced geometry and original attributes to achieve accurate attribute recoloring through the detail-aware K-Nearest Neighbors (DA-KNN) algorithm at the encoder side. Because the enhanced geometry highly preserves the original structure, PAE can effectively retain high-frequency attribute details that are consistent with the original point cloud for efficient attribute compression. After obtaining the reconstructed attribute at the decoder side, we introduce PoAE, which uses a U-Net-based network trained with a weighted mean squared error (W-MSE) loss function to focus on reconstructing attribute residuals, especially in high-frequency regions. During training, PoGE uses a binary cross-entropy (BCE) loss to supervise voxel occupancy, where labels are derived from the binary occupancy maps of the original point cloud. PoAE uses the proposed W-MSE loss to guide the network.

In summary, the main contributions of this paper are as follows:
\begin{itemize}
    \item We propose UGAE, a unified framework for geometry and attribute enhancement in point cloud compression. UGAE simultaneously addresses distortions in both geometry and attributes through three  components: PoGE, PAE, and PoAE.
    \item PoGE uses Transformer blocks and a U-Net architecture to effectively extract both local and global multi-scale geometry features. These features are further enhanced by dense connections in the geometry enhancement head, which fuse information from different channels.
    \item PAE is introduced at the encoder side, to improve attribute quality by combining the enhanced geometry with the original attributes through DA-KNN recoloring before compression.
    \item PoAE is designed for post-attribute enhancement, with the W-MSE loss function guiding the network to focus on quality improvement, especially in high-frequency regions.
    \item UGAE significantly improved geometry and attribute quality in both objective metrics and subjective visual perception on three commonly used datasets: 8iVFB, Owlii, and MVUB.
\end{itemize}

The remainder of this paper is organized as follows. In Section~\ref{sec:related_work}, we review related work in the areas of  point cloud compression, point cloud geometry enhancement, and point cloud attribute enhancement. Section~\ref{sec:method} provides a detailed description of our method, including the methodology of the proposed network and the structure of each module. In Section~\ref{sec:experiments}, we present experimental results and an ablation study to demonstrate the effectiveness of our method. Finally, Section~\ref{sec:conclusion} concludes the paper.

\section{Related Work}
\label{sec:related_work}
\subsection{Point Cloud Compression}
In traditional point cloud compression, two major technologies are predominantly used: G-PCC and V-PCC. G-PCC uses an octree structure to recursively divide the 3D space into sub-cubes (i.e., nodes), encoding the geometry layer by layer. Each node is encoded using an adaptive arithmetic encoder with context models that are manually designed. Following this, attribute compression relies on the reconstructed geometry and selects one of the following methods for attribute prediction: Region-Adaptive Hierarchical Transform (RAHT) \cite{ref50}, Predictive Transform, or Lifting Transform \cite{ref3}. The predicted attributes are then quantized and encoded using arithmetic coding. On the other hand, V-PCC projects the 3D point cloud into 2D geometry and attribute videos, which are subsequently compressed using video coding standards such as H.265/HEVC.

Learning-based point cloud compression methods also sequentially compress geometry and attributes. In geometry compression, research primarily focuses on four categories of methods: point-based methods \cite{ref7, ref8, ref9, ref10, ref11, ref12} use network architectures such as PointNet \cite{ref51} and PointNet++ \cite{ref52} to build autoencoder structures, applying Chamfer Distance (CD) loss for supervision to compress point clouds; voxel-based methods \cite{ref13, ref14, ref15, ref16} rely on dense convolutions to construct autoencoders, with loss functions typically being Binary Cross-Entropy (BCE) or Generalized Focal Loss \cite{ref53}; octree-based methods \cite{ref17, ref18, ref19, ref20, ref21, ref22, ref23, ref24} apply deep learning to improve octree context modeling by exploring correlations among parent, child, and sibling nodes to more accurately estimate probabilities; and sparse tensor-based methods \cite{ref25, ref26, ref27, ref28} use efficient sparse convolutions to build autoencoders. Among these four categories of geometry compression methods, point-based methods are mostly used for lossy compression, while the other three categories can be applied to both lossy and lossless compression. For attributes, compared to lossless compression \cite{ref29, ref30, ref31, ref32}, lossy compression \cite{ref33, ref34, ref35, ref36, ref54} achieves much lower bitrates while maintaining high perceptual quality. Notable methods include Deep PCAC \cite{ref33} and Sparse PCAC \cite{ref34}, where the former uses 3D convolutions and the latter applies sparse convolutions. Additionally, SPAC \cite{ref55} introduces frequency band splitting and achieves superior R-D performance compared to G-PCC.

\subsection{Geometry Enhancement}
Lossy geometry compression often leads to point disappearance issue. Yu et al. \cite{ref37} proposed PU-Net, one of the earliest deep learning methods designed to address this problem. It uses PointNet++ \cite{ref52} to extract multi-scale features. Qian et al. proposed PUGeo-Net \cite{ref56}, which achieves efficient upsampling by learning local geometric parameters and normal vectors for each point, sampling in the 2D parameter domain, and combining learned 3D geometric transformations. PU-GAN \cite{ref57} introduces generative adversarial networks (GAN) for point cloud upsampling. Liu et al. proposed a frequency-aware upsampling network PUFA-GAN \cite{ref38} that not only generates dense point clouds on the underlying surface but also effectively suppresses high-frequency noise. Liu et al. also proposed PU-Mask \cite{ref39} which introduces a virtual mask mechanism to guide point cloud upsampling, aiming at filling locally sparse regions. However, these point-based methods typically upsample by dividing the point cloud into smaller patches, which limits their ability to learn from global context. PU-Dense \cite{ref40} was the first to address large-scale point cloud upsampling by introducing multi-scale sparse convolutions. It uses a binary voxel occupancy classification loss to train the network, enabling efficient processing of high-resolution point clouds with millions of points. For G-PCC compressed data, Fan et al. \cite{ref58} proposed DGPP, which extracts features using 3D convolutions and estimates occupancy through a multi-scale probabilistic prediction mechanism from coarse to fine granularity. Ding et al. \cite{ref41} analyzed the main causes of geometry distortion in different types of compressed point clouds and design a module selection strategy that adaptively chooses repair modules based on auxiliary information.

\subsection{Attribute Enhancement}
To address the artifacts caused by attribute compression, traditional methods introduce various filtering techniques \cite{ref59, ref60, ref61, ref62}. Wang et al. \cite{ref59} introduced a Kalman filter into the G-PCC framework to optimize the reconstructed attributes. Subsequently, improved methods based on Wiener filtering \cite{ref60, ref61, ref62} were proposed, which can effectively alleviate distortion accumulation during the encoding process, thereby further improving reconstruction quality. However, these methods are still limited by the assumption of linear distortion modeling.

In contrast, deep learning methods proposed in recent years have demonstrated stronger modeling capabilities in attribute restoration. Existing deep learning approaches can be broadly categorized into the following three categories: graph convolution-based methods \cite{ref46}, \cite{ref63}, sparse convolutions-based methods \cite{ref64, ref65, ref66}, and projection-based methods \cite{ref44}, \cite{ref67}. In the domain of graph convolution, MS-GAT \cite{ref63} is the first work targeting attribute compression distortions in G-PCC. It proposes a multi-scale graph attention network, where the decoded geometry coordinates are used to construct the graph structure, and the compressed attributes serve as vertex signals. Xing et al. proposed GQE-Net \cite{ref46} that further incorporates normal vectors and geometry distances as auxiliary information. However, these graph-based methods typically partition point clouds into small patches of fixed size (e.g., 2048 points), leading to high inference costs and limited scalability when applied to large-scale point clouds. In contrast, sparse convolutions offer advantages in computational efficiency and memory usage. Liu et al. proposed DAE-MP \cite{ref64}, a post-processing enhancement method that uses sparse convolutions for dynamic point cloud compression, achieving improved quality through explicit motion estimation and frequency-aware processing. Ding et al. proposed CARNet \cite{ref65}, an adaptive loop filtering network based on dual-stream sparse convolutions and dynamic linear weighting. Subsequently, Zhang et al. \cite{ref66} introduced two solutions based on sparse convolutions for G-PCC compressed point cloud attributes: NeuralSAO and NeuralBF. Besides direct 3D point cloud processing, some methods adopt projection-based strategies. OCARNet \cite{ref67} leverages occupancy information as prior knowledge to guide the network in focusing on attribute distortions within occupied regions, effectively removing compression artifacts from V-PCC decoded attribute images. Xing et al. \cite{ref44} proposed SSIU-Net which first transforms 3D point cloud patches into 2D images, enhances them using a lightweight U-Net variant, and then maps the enhanced 2D results back to the corresponding 3D patches.

The aforementioned methods primarily address either geometry or attribute distortions caused by compression. In practical scenarios, however, geometry and attributes are often compressed together, and enhancing only one aspect may be insufficient to fully enhance overall point cloud quality. To the best of our knowledge, G-PCC++ \cite{ref42} is the only approach that considers geometry and attribute distortions in a unified manner. However, this method only carries out point cloud enhancement at the decoder side. It first enhances the lossy geometry, then interpolates the attributes based on the enhanced geometry, and finally enhances the attributes. Consequently, it fails to fully exploit the potential benefits that geometry enhancement can bring to attribute compression and reconstruction. Moreover, re-coloring based on lossy geometry at the encoder side may further degrade the original attribute information. To address these challenges, we propose the UGAE framework, which takes a systematic approach to jointly alleviate the distortions in both geometry and attribute compression. By integrating geometry enhancement with attribute reconstruction, UGAE significantly improves the overall reconstruction quality of compressed point clouds.

\section{Proposed Method}
\label{sec:method}
\subsection{Problem Statement}
Given an original point cloud $\mathbf{P}=\{\mathbf{G},\mathbf{A}\}$, where $\mathbf{G}=(\bm{\mathit{x}},\bm{\mathit{y}},\bm{\mathit{z}})\in\mathbb{R}^{N\times3}$ denotes the 3D geometry coordinates and $\mathbf{A}=(\bm{\mathit{r}},\bm{\mathit{g}},\bm{\mathit{b}})\in\mathbb{R}^{N\times3}$ represents the attribute information, with $N$ being the number of points, existing lossy compression methods (e.g., G-PCC, V-PCC) achieve data reduction through quantization and entropy coding. However, these techniques inevitably introduce distortions in both geometry and attributes. Geometry distortion leads to significant structural differences between the compressed geometry ${\mathbf{\bar G}}$ and the original $\mathbf{G}$. Attribute distortion, on the other hand, often suppresses high-frequency details (e.g., textures and boundaries) due to quantization, resulting in artifacts and color shifts in the reconstructed attributes $\mathbf{\bar A}$ and such distortion is often irreversible. Most existing works optimize geometry or attributes separately, neglecting their coupled relationship. Even joint enhancement method \cite{ref42} fail to explicitly model how geometry enhancement can guide the recovery of high-frequency attribute details. Moreover, due to the blurred structure in compressed geometry $\mathbf{\bar G}$, re-coloring based on it leads to excessive loss of high-frequency details; even if interpolation and enhancement of decoded attributes are applied using the enhanced geometry at the decoder, it remains difficult to recover the missing high-frequency details. To address these challenges, we propose the UGAE framework.
\begin{figure*}[!t]
\centering
\includegraphics[width=0.9\linewidth]{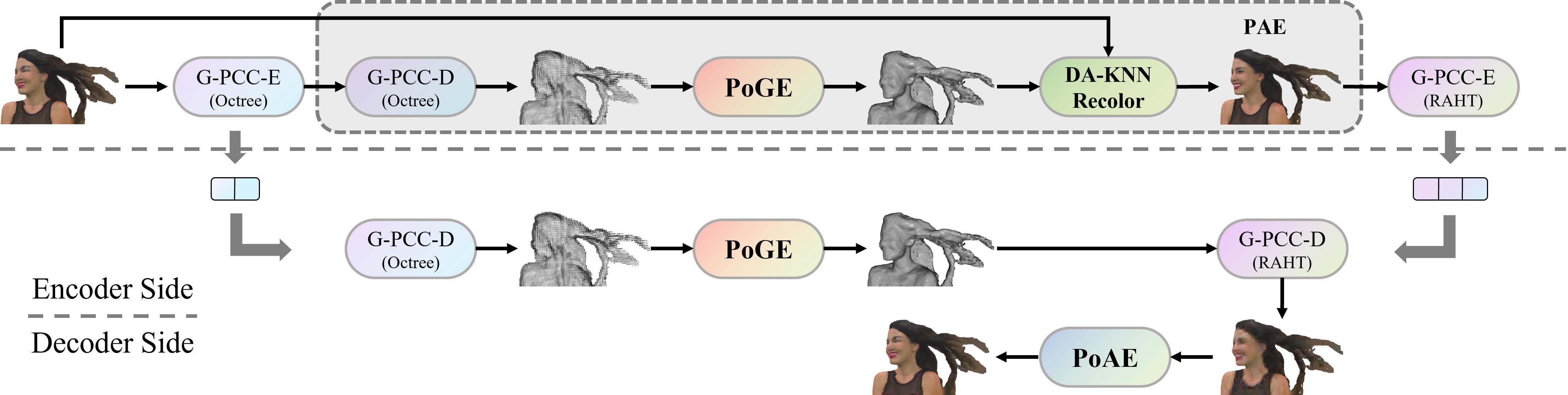}
\caption{UGAE pipeline. At the encoder side, PoGE enhances the lossy geometry, and PAE recolors the enhanced geometry using the original attribute information. At the decoder side, PoGE reconstructs the same enhanced geometry as in the encoder side to assist in attribute decoding, and PoAE focuses on high-frequency regions to produce the final enhanced point cloud.}
\label{fig:UGAE_arch}
\end{figure*}

\subsection{UGAE Framework}
The overall architecture of UGAE based on the G-PCC compression pipeline is illustrated in Fig.~\ref{fig:UGAE_arch}. UGAE consists of three key parts. At the encoder side, PoGE is first used to restore an enhanced geometry $\hat{\mathbf{G}}$ that aligns with the original structure. Then, using the original attributes $\mathbf{A}$, PAE applies DA-KNN recoloring to $\hat{\mathbf{G}}$ to generate an intermediate attribute $\widetilde{\mathbf{A}}$, allowing the intermediate point cloud $\widetilde{\mathbf{P}}=\{\hat{\mathbf{G}},\widetilde{\mathbf{A}}\}$ to preserve more high-frequency details for subsequent attribute compression. As the same enhanced geometry $\hat{\mathbf{G}}$ can be reconstructed at the decoder, attribute-lossy compression can be applied to $\widetilde{\mathbf{P}}$, resulting in the reconstructed point cloud $\ddot{\mathbf{P}}=\{\hat{\mathbf{G}},\mathbf{\bar A}\}$. At the decoder side, PoAE is proposed to further improve attribute quality by compensating for high-frequency regions in the reconstructed attribute $\mathbf{\bar A}$. This results in the final jointly enhanced point cloud $\hat{\mathbf{P}}=\{\hat{\mathbf{G}},\hat{\mathbf{A}}\}$, in which both geometry and attributes are efficiently enhanced.

\subsection{PoGE}
\begin{figure*}[!t]
\centering
\includegraphics[width=\linewidth]{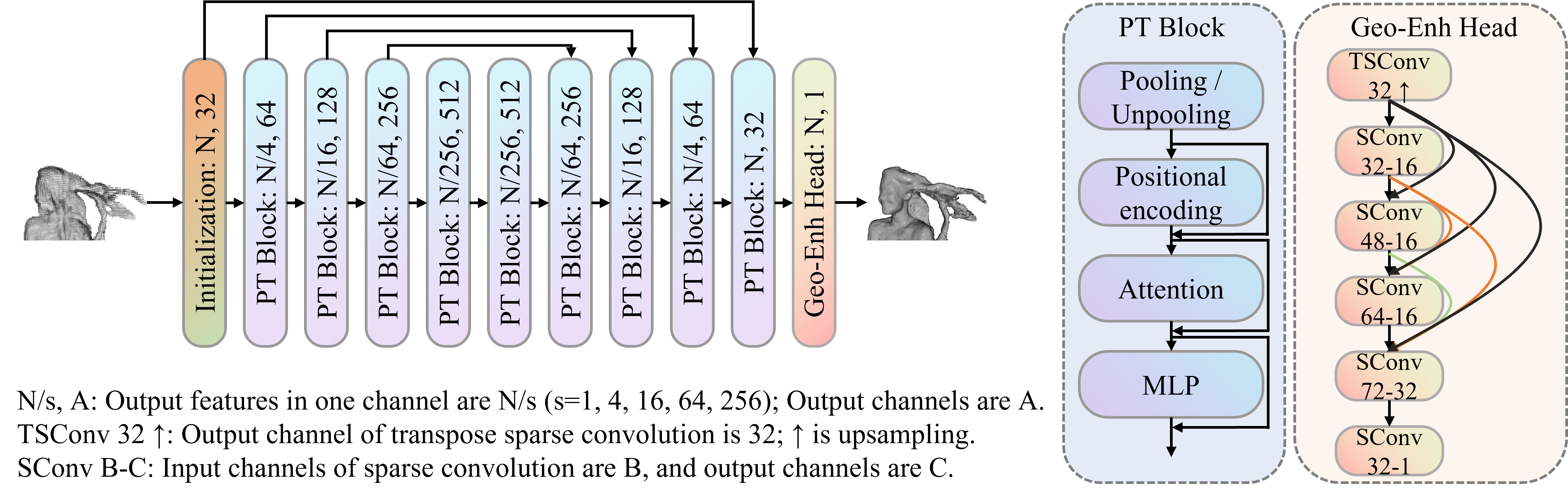}
\caption{PoGE architecture. The initialization converts the unstructured point cloud into a structured format, and PT blocks extract multi-scale features. The enhanced geometry is obtained from the geometry enhancement head (Geo-Enh Head) through probability-based sorting and selection.}
\label{fig:poge}
\end{figure*}

Previous studies have demonstrated that local features contribute to detail recovery, while global features help maintain structural integrity \cite{ref39}, \cite{ref46}, \cite{ref54}. Point Transformer v3 (PT) \cite{ref48} extends the range of self-attention up to 4096 points via a serialization mechanism, significantly outperforming existing methods \cite{ref41}, \cite{ref46}, \cite{ref54}. Based on this, we stack multiple PT blocks to model long-range correlations. To fully capture global features, PoGE uses a multi-scale U-Net architecture. As shown in Fig.~\ref{fig:poge}, PoGE consists of a feature extraction and a geometry enhancement head. In the feature extraction, the Initialization block, motivated by requirement of PT blocks for serialized input, transforms the unstructured point cloud into a structured format. Then, PT blocks progressively downsample the points to $N/256$, extracting high-level semantic information. Multi-scale geometry features, consisting of local details and global structure, are then fused through layer-wise upsampling with skip connections.

As illustrated on the right part of Fig.~\ref{fig:poge}, the geometry enhancement head consists of six sparse convolutional (SConv) layers used to predict the occupancy probability of each point. First, TSConv generates all potential occupied positions. To prevent excessive high memory usage, we introduce a dense connection structure after TSConv. This design not only simplifies the network structure but also facilitates multi-dimension feature fusion aimed at preserving geometry information. Finally, an SConv layer reduces the features into one dimension to output the occupancy probability for each candidate point. The points with the higher probabilities are selected to form the enhanced geometry point set $\hat{\mathbf{G}}$.

Given that voxel occupancy states are binary (occupied or unoccupied), we use a BCE loss function to supervise the difference between the network output and the ground truth during the training of PoGE. The optimization objective is defined as
\begin{equation}
    \min_{\theta_{\mathrm{PoGE}}} \mathrm{BCE}\left(f_{\mathrm{PoGE}}(\mathbf{G}; \theta_{\mathrm{PoGE}}), \mathbf{M}_{\mathrm{occ}}\right),
\end{equation}
where $f_{\mathrm{PoGE}}(\bullet)$ represents PoGE, which outputs voxel occupancy probabilities, $\theta_{\mathrm{PoGE}}$ denotes the network parameters, and $\mathbf{M}_{\mathrm{occ}} \in \{0,1\}^n$ is the voxel occupancy mask of the ground truth point cloud.

\begin{figure}
\centering
\includegraphics[width=\linewidth]{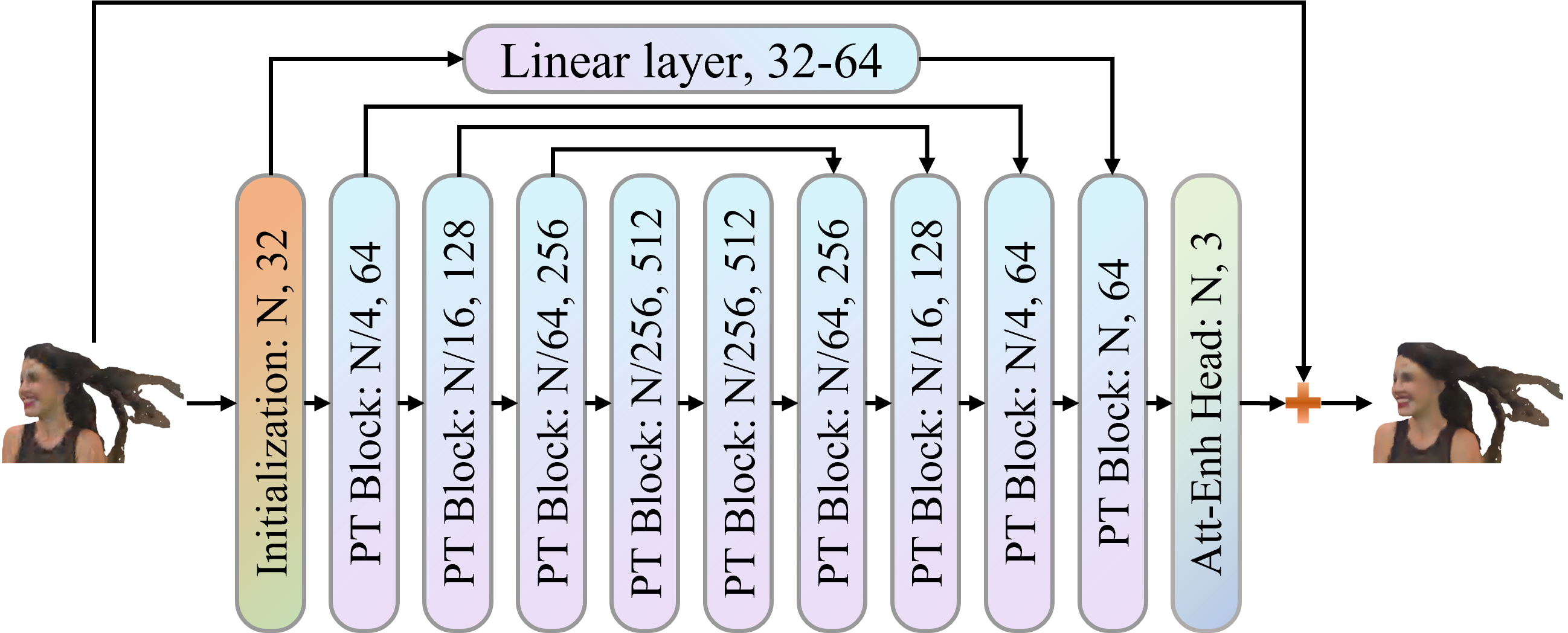}
\caption{PoAE architecture. PT blocks extract multi-scale features, and the attribute enhancement head (Att-Enh Head) predicts color residuals, which are added to the original attributes to obtain the enhanced attributes.}
\label{fig:poae}
\end{figure}

\begin{figure}[ht]
\centering
\includegraphics[width=\linewidth]{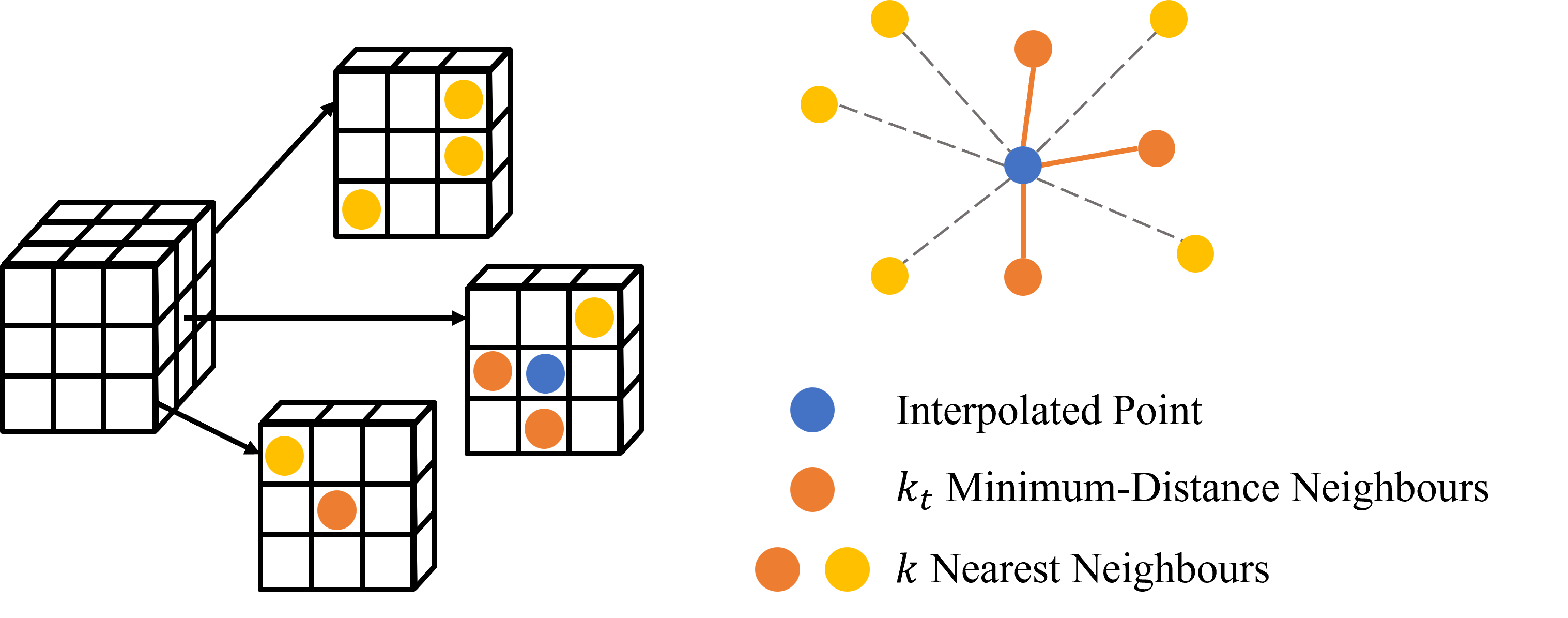}
\caption{Illustration of DA-KNN recoloring for $k=8$. The left side shows a query point (shown in blue) in the enhanced geometry along with its neighboring points (shown in yellow and orange); the right side illustrates the spatial distribution of neighbors in 3D space. DA-KNN first finds the $k$ nearest neighbours (shown in yellow and orange) of the query point (shown in blue). Then, from these $k$ neighbours, the algorithm selects the closest ones that lie at the same distance from the query point. In this example, $k_t=3$ points (shown in orange) are selected. }
\label{fig:DAKNN}
\end{figure}

During our experiments, we observed that even after fixing all explicit sources of randomness (e.g., network initialization, shuffle orders in PTv3 blocks), it was still impossible to obtain the same geometry enhancement results across multiple test runs. After extensive debugging and controlled comparisons, we identified that the remaining uncertainty mainly stems from the non-deterministic behavior of TSConv. TSConv leverages GPU-based parallel computing to accelerate inference. However, this introduces the following issue: the order in which points are fed into the GPU may vary across different inference passes. Moreover, the non-deterministic execution order of ``atomic'' operations of GPU results in inconsistent feature calculation orders. To address this problem, we explored two potential solutions:
\begin{itemize}
\item \textbf{Decimal Shifting}: Before feeding data into TSConv, the fractional part of each floating-point is shifted to the integer part, and then restored afterward. However, due to the floating-point limitations of the sparse convolutions, the shifting may result in truncation of trailing digits, causing information loss.

\item \textbf{CPU-based TSConv Inference}: Running TSConv on the CPU ensures serial execution, which guarantees deterministic outputs. 
\end{itemize}

Considering the trade-off between consistency and performance, we opted for the second solution---deploying TSConv on the CPU during inference to ensure reproducibility of the enhanced geometry.

\subsection{PAE}
We migrate the TSConv layer to the CPU to ensure result reproducibility. This design allows both the encoder side and decoder side to obtain the same enhanced geometry $\hat{\mathbf{G}}$ from the same lossy geometry $\mathbf{\bar G}$. Leveraging this advantage, we recolor the original point cloud attributes $\mathbf{A}$ onto the enhanced geometry $\hat{\mathbf{G}}$, generating a new point cloud for attribute compression $\widetilde{\mathbf{P}} = \{\hat{\mathbf{G}}, \widetilde{\mathbf{A}}\}$, as shown in Fig.~\ref{fig:UGAE_arch}. Given that the enhanced geometry structurally retains the primary information of the original geometry, re-coloring yields attributes $\widetilde{\mathbf{A}}$ that remain close to the original attributes $\mathbf{A}$, preserving much of the high-frequency details and facilitating enhancement after decoding. Because of the reproducibility of enhanced geometry, only the lossy geometry bitstream and the lossy recolored attribute bitstream need to be transmitted.

A core component in PAE is attribute recoloring. Compared to the lossy geometry $\mathbf{\bar G}$, the enhanced geometry $\hat{\mathbf{G}}$ exhibits a denser point distribution. Therefore, weighted summation-based recoloring (e.g., using all neighbors with weighted averaging) performs poorly in this task, as it may incorporate attributes from distant points and leads to the loss of high-frequency details. To adapt to the increased geometry density and preserve high-frequency details as much as possible, we propose the DA-KNN algorithm. Given the uniform spacing between voxels in the voxelized point cloud, many neighboring points may lie at equal distances from a query point. Based on this observation, for each query point in the enhanced geometry $\hat{\mathbf{G}}$, we first identify its $k$ nearest neighbours from the original point cloud. Among these, we then select those that are closest to the query point and equidistant from it (Fig.~\ref{fig:DAKNN}). These $k_t$ neighbors form a local neighborhood $\{{g_i}^l, {a_i}^l\}_{l=1}^{k_t}$, used to interpolate the attribute of the query point as follows
\begin{equation}
    \widetilde{a}_i = \frac{1}{k_t} \sum_{l=1}^{k_t} a_i^l.
\end{equation}

Because attributes typically vary smoothly within a local region, averaging the attributes of these nearby neighbors does not significantly blur high-frequency details.

\subsection{PoAE}
\begin{figure}[!t]
\centering
\includegraphics[width=0.9\linewidth]{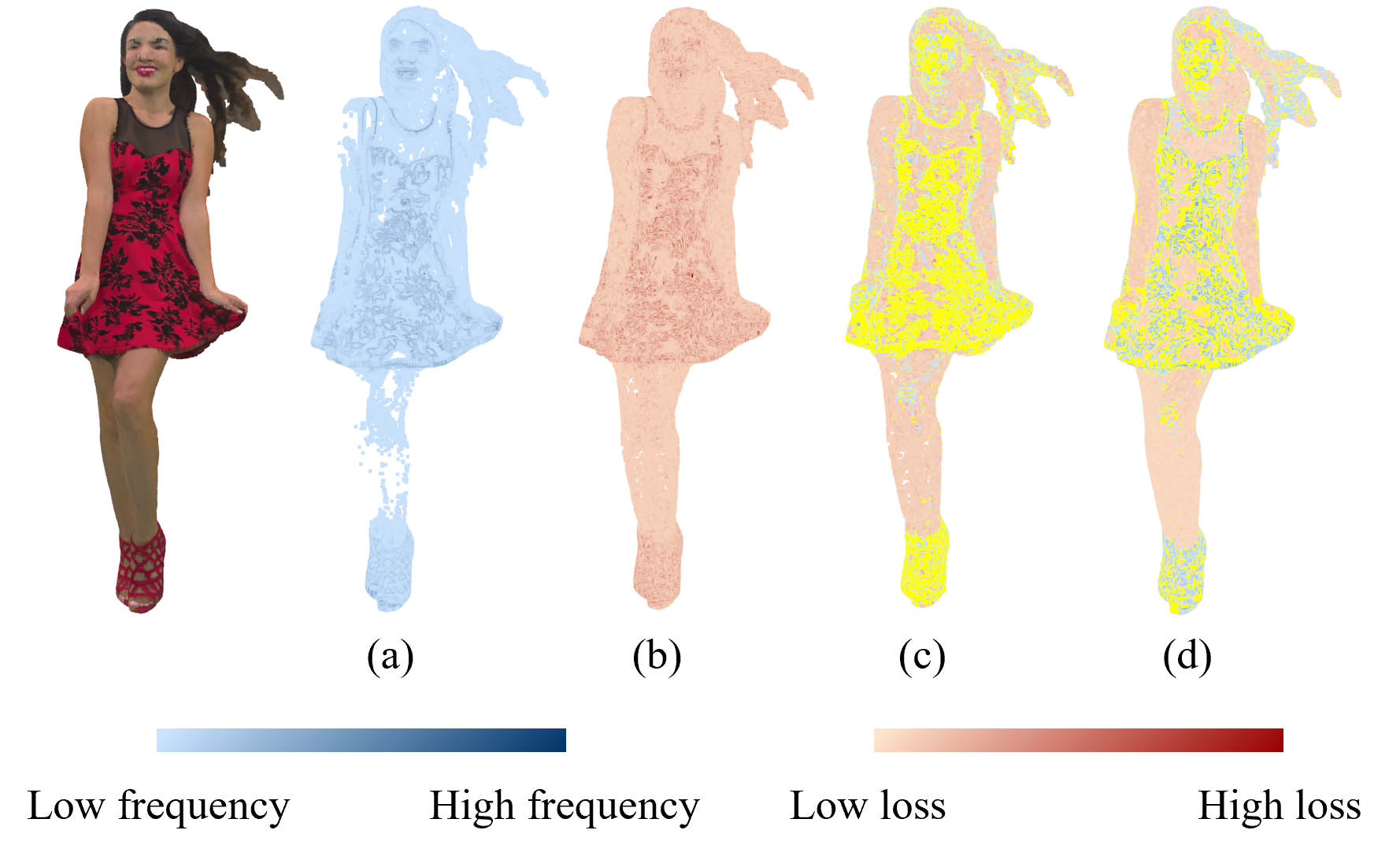}
\caption{(a) High-frequency (top 50\%) regions in the original point cloud attributes. 
        (b) High-loss (top 50\%) regions in the decoded point cloud attributes. 
        (c) Overlap between (a) and (b): yellow indicates overlapping areas, while blue and light red denote non-overlapping regions; the overlap ratio is 75.09\%. 
        (d) Overlap visualization after PoAE, with an overlap ratio of 53.08\%.}
\label{fig:high_low_regions}
\end{figure}

As shown in Fig.~\ref{fig:high_low_regions}, we distinguish high-frequency and low-frequency regions by measuring the color differences between each point and its neighboring points. Additionally, high-loss and low-loss regions are distinguished by evaluating the differences between the reconstructed attributes and the original attributes. The most significant attribute distortions in the decoded point cloud primarily lie in the high-frequency regions of the original attributes, accounting for up to 75.09\%. To improve attribute quality, especially by alleviating the loss of high-frequency details, we use PoAE after decoding.

The structure of PoAE is similar to PoGE, except that in the last block, the extracted feature dimension from initialization is increased to 64 via a linear layer before being fused with the output from the penultimate block, as shown in Fig.~\ref{fig:poae}. After multi-scale feature extraction, the attribute enhancement head (a single linear layer) reduces the features to three dimensions to predict residuals for the RGB channels. These residuals are then added to the initially enhanced attributes $\mathbf{\bar A}$ to obtain the final enhanced attributes $\hat{\mathbf{A}}$.

As regions with high attribute loss are strongly correlated with high-frequency regions, we assign greater loss weights to these areas to better supervise the network. While the standard MSE loss naturally penalizes large deviations, it does not sufficiently emphasize regions with high loss. To address this issue, we design a high-loss-aware weighted W-MSE loss function. For the enhanced attributes $\hat{\mathbf{A}}$ from PoAE, we compute the W-MSE loss with the recolored attributes $\widetilde{\mathbf{A}}$:
\begin{equation}
    \mathcal{L}_{\mathrm{W-MSE}} = \frac{1}{N} \sum_{i=1}^{N} w_i \cdot (\hat{a}_i - \widetilde{a}_i)^2,
\end{equation}
where $\hat{a}_i$ is the $i$-th enhanced attribute, $\widetilde{a}_i$ is the $i$-th recolored attribute, and the weight $w_i$ for each sample is defined as
\begin{equation}
    w_i =
    \begin{cases}
        w_{\mathrm{high}}, & \text{if } e_i > T, \\
        w_{\mathrm{low}},  & \text{otherwise},
    \end{cases}
\end{equation}
and the threshold $T = \mathrm{Quantile}(\{e_i\}_{i=1}^N, q)$ is determined based on the $q$-th quantile of all errors, with $e_i = |\hat{a}_i - \widetilde{a}_i|$, $i = 1, 2, \ldots, N$.

After enhancement via PoAE, the attribute loss in high-frequency regions of the point cloud $\hat{\mathbf{P}} = \{\hat{\mathbf{G}}, \hat{\mathbf{A}}\}$ is significantly reduced. As shown in Fig.~\ref{fig:high_low_regions}(d), high loss regions have shifted from being concentrated in the high-frequency regions to being more prominent in the low-frequency ones. Furthermore, the overlap between high loss regions in the enhanced point cloud and the original high-frequency regions reached 53.08\%, which illustrates the effectiveness of PoAE in enhancing high-frequency details.
\begin{table}[t]
    \centering
    \caption{Correspondence Between Rate Levels, PQS, and QP for G-PCC Compression}
    \label{tab:compression_params}
    \begin{tabular}{cccccc}
        \hline
        Rate Level & R01   & R02   & R03  & R04  & R05 \\
        \hline
        PQS        & 0.125 & 0.25  & 0.5  & 0.75 & 0.875 \\
        QP         & 51     & 46    & 40   & 34   & 28 \\
        \hline
    \end{tabular}
\end{table}

\section{Results and Analysis}
\label{sec:experiments}
We conducted extensive quantitative and qualitative experiments to thoroughly evaluate the performance of the proposed UGAE under lossy compression of both geometry and attributes. We also compared the proposed framework with existing state-of-the-art enhancement approaches. In addition, we carried out multiple ablation studies to analyze the contribution of each UGAE component to the overall performance.
\subsection{Datasets}
\subsubsection{Training datasets}
\begin{figure}
\centering
\includegraphics[width=\linewidth]{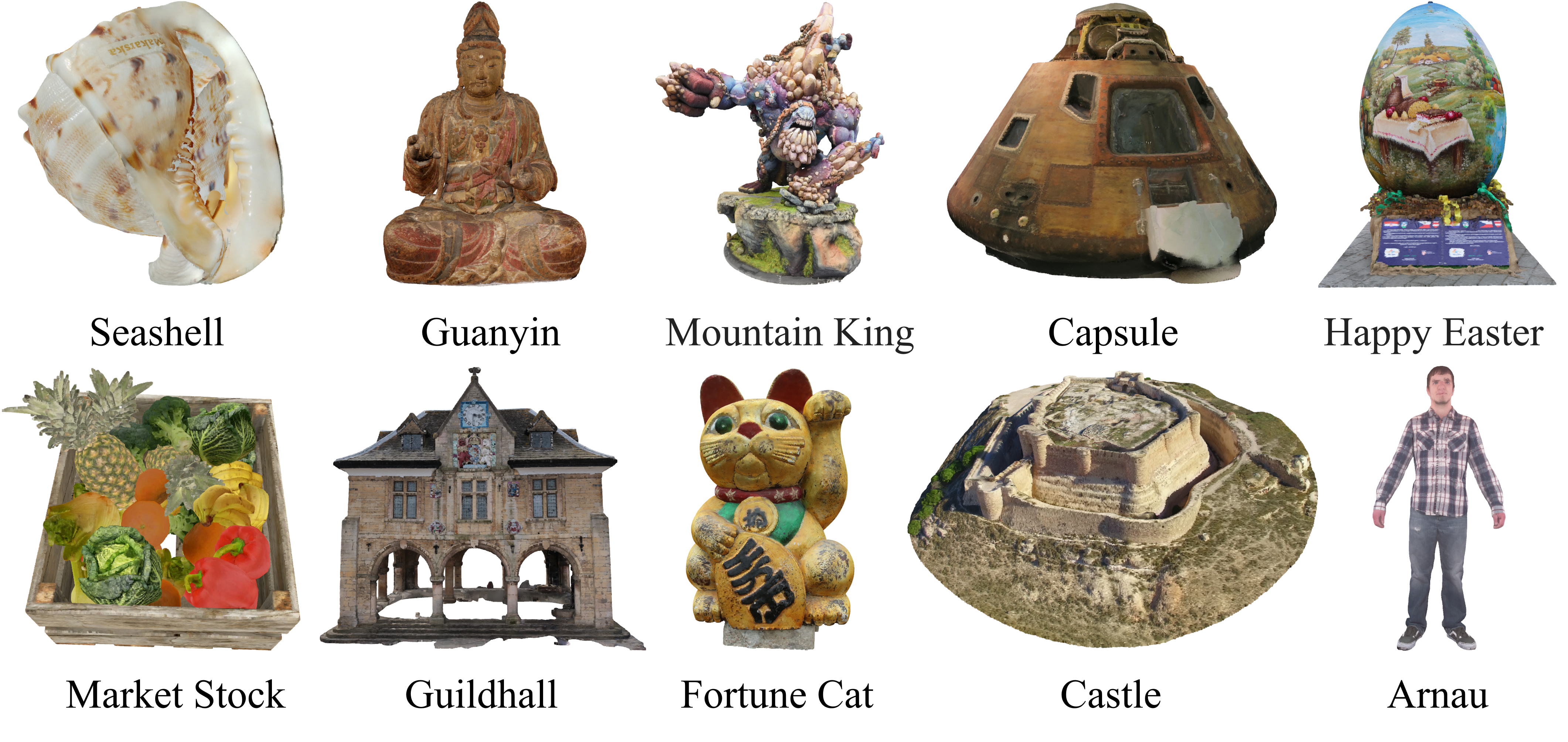}
\caption{Sample examples from the training dataset.}
\label{fig:dataset_partition}
\end{figure}

We used the Real-World Textured Things (RWTT) dataset~\cite{ref68} to train the proposed UGAE framework. From each 3D model, we extracted up to $2 \times 10^6$ points and voxelized them to a resolution of $10^3$. We then applied the KDTree method to partition each voxelized 3D model (as shown in Fig.~\ref{fig:dataset_partition}) into sub-point clouds, each containing no more than 100,000 points. Finally, we obtained 8,510 sub-point clouds, with the first 1,000 reserved for validation and the remaining 7,510 used for training.

Subsequently, we used the Octree + RAHT configuration of G-PCC Test Model Category 13 version 29.0 (TMC13v29) to compress these sub-point clouds, adhering strictly to the Common Test Conditions (CTC) of G-PCC. Table~\ref{tab:compression_params} details the key parameter settings used during compression, including the geometry quantization parameter \emph{Position Quantization Scale} (PQS) and the attribute quantization parameter \emph{Quantization Parameter} (QP). It is important to note that PoGE was trained using the original geometry $\mathbf{G}$ and lossy geometry ${\mathbf{\bar G}}$, whereas PoAE was trained using recolored point clouds $\widetilde{\mathbf{P}}$ and reconstructed point clouds $\ddot{\mathbf{P}}$.

\subsubsection{Test datasets}
To evaluate the generalization capability and practical performance of UGAE, we selected three publicly available test datasets that are widely used in the field of point cloud compression: 8i Voxelized Full Bodies (8iVFB)~\cite{ref69}, Owlii~\cite{ref70}, and Microsoft Voxelized Upper Bodies (MVUB)~\cite{ref71}.

\begin{table*}[t]
\centering
\caption{Quantitative gains of UGAE compared to G-PCC (TMC 13v29) in terms of various metrics.}
\label{tab:results}
\scriptsize
\resizebox{\textwidth}{!}{
\begin{tabular}{l*{6}{cc}}
\toprule
\multirow{3}{*}{Point Cloud} & \multicolumn{2}{c}{D1} & \multicolumn{2}{c}{D2} & \multicolumn{2}{c}{Y} & \multicolumn{2}{c}{YUV} & \multicolumn{2}{c}{1-PCQM} & \multicolumn{2}{c}{IWSSIM\textsubscript{p}} \\
\cmidrule(lr){2-3} \cmidrule(lr){4-5} \cmidrule(lr){6-7} \cmidrule(lr){8-9} \cmidrule(lr){10-11} \cmidrule(lr){12-13}
& BD-BR & BD-PSNR & BD-BR & BD-PSNR & BD-BR & BD-PSNR & BD-BR & BD-PSNR & BD-BR & BD-PCQM & BD-BR & BD-IWSSIM\textsubscript{p} \\
& (\%) & (dB) & (\%) & (dB) & (\%) & (dB) & (\%) & (dB) & (\%) & ($10^{-3}$) & (\%) & ($10^{-2}$) \\
\midrule
longdress     & -88.71 & 9.44  & -77.14 & 7.33  & -49.74 & 2.49 & -49.00 & 2.45 & -66.78 & 12.64 & -71.53 & 0.1465 \\
loot          & -90.43 & 10.72 & -82.32 & 8.81  & -49.50 & 2.99 & -50.31 & 3.00 & -70.22 & 12.30 & -78.01 & 0.1660 \\
redandblack   & -88.23 & 9.55  & -79.00 & 7.73  & -52.74 & 2.99 & -51.49 & 2.89 & -69.04 & 12.00 & -75.74 & 0.1458 \\
soldier       & -88.58 & 9.91  & -78.56 & 8.03  & -47.95 & 2.86 & -47.27 & 2.69 & -68.71 & 16.12 & -70.56 & 0.1748 \\
\midrule
\textbf{Average} & \textbf{-88.99} & \textbf{9.91} & \textbf{-79.26} & \textbf{7.98} & \textbf{-49.99} & \textbf{2.83} & \textbf{-49.52} & \textbf{2.76} & \textbf{-68.69} & \textbf{13.26} & \textbf{-73.96} & \textbf{0.1583} \\
\midrule
basketball    & -93.02 & 11.75 & -87.29 & 10.11 & -64.55 & 4.24 & -63.51 & 3.89 & -73.97 & 9.19 & -79.74 & 0.1843 \\
dancer        & -92.35 & 11.61 & -85.24 & 9.55  & -64.38 & 4.46 & -63.42 & 4.14 & -72.46 & 10.08 & -79.49 & 0.1848 \\
exercise      & -92.43 & 11.62 & -85.84 & 9.76  & -57.27 & 3.11 & -56.52 & 2.92 & -71.33 & 8.26 & -79.21 & 0.1832 \\
model         & -92.24 & 11.18 & -84.29 & 8.82  & -64.78 & 4.26 & -64.55 & 4.01 & -75.76 & 12.01 & -81.67 & 0.1990 \\
\midrule
\textbf{Average} & \textbf{-92.51} & \textbf{11.54} & \textbf{-85.67} & \textbf{9.56} & \textbf{-62.75} & \textbf{4.02} & \textbf{-62.00} & \textbf{3.74} & \textbf{-73.38} & \textbf{9.88} & \textbf{-80.03} & \textbf{0.1878} \\
\midrule
andrew        & -90.23 & 7.93  & -76.26 & 5.50  & -56.37 & 2.91 & -55.27 & 2.63 & -69.90 & 11.05 & -78.76 & 0.1586 \\
david         & -92.19 & 9.29  & -83.98 & 7.32  & -55.68 & 3.93 & -55.63 & 3.77 & -60.35 & 8.24 & -80.16 & 0.1942 \\
phil          & -90.86 & 8.54  & -80.06 & 6.48  & -67.49 & 4.58 & -66.76 & 4.24 & -77.20 & 16.23 & -82.60 & 0.1940 \\
ricardo       & -91.74 & 8.98  & -80.52 & 6.50  & -48.54 & 3.82 & -48.71 & 3.66 & -42.56 & 4.41 & -89.50 & 0.2056 \\
sarah         & -91.74 & 9.24  & -82.43 & 7.06  & -60.37 & 5.10 & -60.46 & 4.91 & -70.22 & 8.28 & -83.37 & 0.1951 \\
\midrule
\textbf{Average} & \textbf{-91.35} & \textbf{8.80} & \textbf{-80.65} & \textbf{6.57} & \textbf{-57.69} & \textbf{4.07} & \textbf{-57.36} & \textbf{3.84} & \textbf{-64.05} & \textbf{9.64} & \textbf{-82.88} & \textbf{0.1895} \\
\bottomrule
\end{tabular}
}
\end{table*}

\begin{figure*}
\centering
\includegraphics[width=\linewidth]{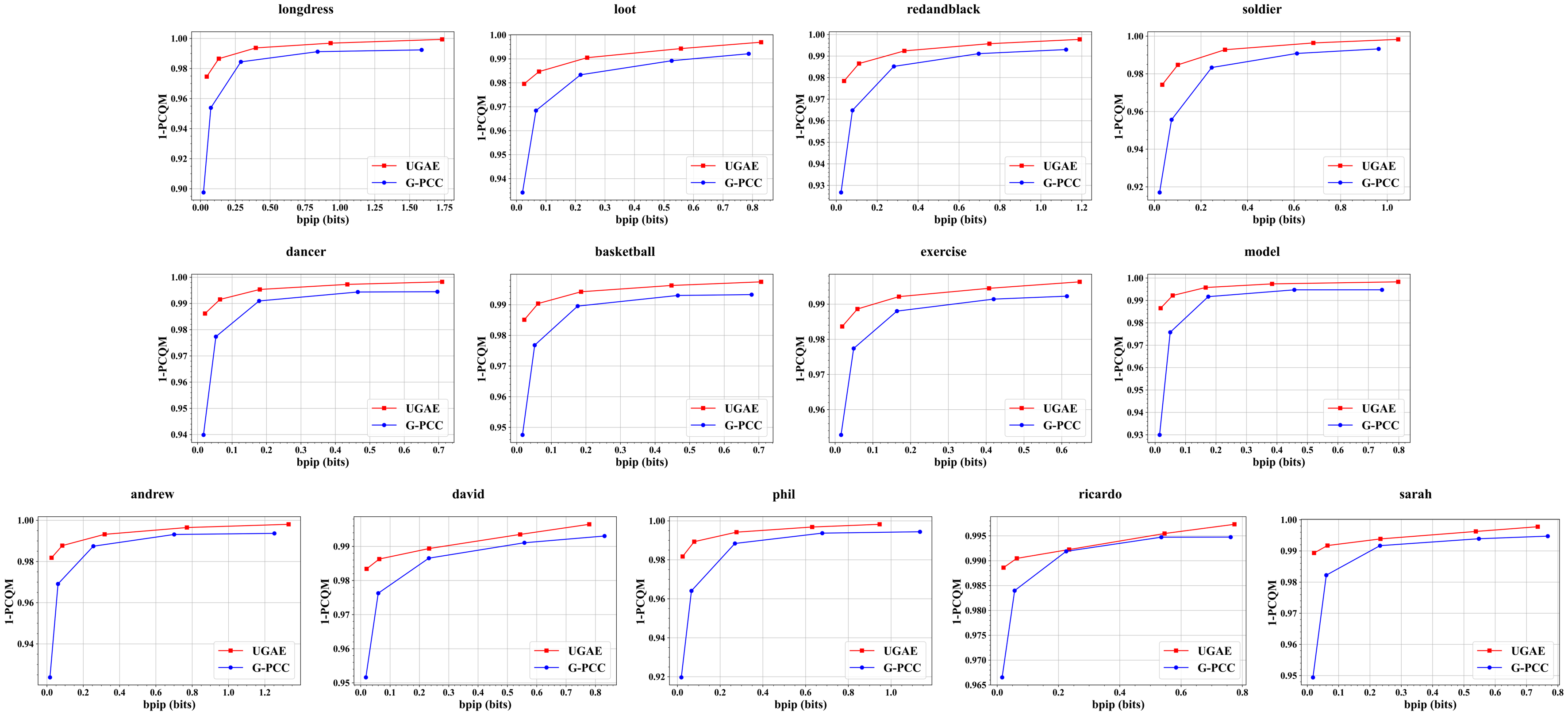}
\caption{R-D curves of all test point clouds.}
\label{fig:rd_curves}
\end{figure*}

\subsection{Implementation Details}
The proposed UGAE was implemented based on PyTorch 2.1.1 and MinkowskiEngine 0.5.4, running on a workstation with an Intel Xeon Gold 6148 CPU, 640~GB RAM, and an Nvidia RTX 4090 GPU. PoGE and PoAE networks were both trained using AdamW for 100 epochs, with learning rates of 0.006 and 0.003, and batch sizes of 2 and 5, respectively. For the recolor operation in PAE, we used DA-KNN with $k = 8$ neighbors. For the W-MSE loss function, we set $w_{\mathrm{high}} = 2$, $w_{\mathrm{low}} = 0.5$, and $q = 0.4$. These values were chosen for simplicity and were not exhaustively tuned. Our goal is to demonstrate the effectiveness of W-MSE rather than to optimize performance through parameter search.

To evaluate the performance of UGAE, we used bits per input point (bpip) as the metric for bitrate. We measured geometry reconstruction quality using point-to-point distance (D1 PSNR) and point-to-plane distance (D2 PSNR). For attribute quality, we used the Y-PSNR metric to evaluate the primary luminance component, and YUV-PSNR~\cite{ref72} with a Y:U:V component ratio of 14:1:1 to assess the overall color quality. To comprehensively reflect visual quality, the PCQM~\cite{ref73} and  IWSSIM\textsubscript{p}~\cite{ref74} metrics were used to objectively evaluate the quality of colored point clouds. To provide a detailed rate-distortion (R-D) performance analysis, we used the Bjøntegaard Delta (BD)~\cite{ref75} metrics\footnote{To compute the BD metrics, we used the code: \url{https://github.com/FAU-LMS/bjontegaard-matlab}}: BD-Bitrate (BD-BR), BD-PSNR, BD-PCQM, and BD-IWSSIM\textsubscript{p}, using Akima interpolation for curve fitting. A negative BD-BR and positive BD-PSNR or BD-PCQM or IWSSIM\textsubscript{p} indicate a positive gain by the proposed method.

\subsection{Objective Quality Evaluation}
Table~\ref{tab:results} presents the objective quality evaluation results of the proposed method on the three standard benchmark datasets in terms of both geometry and attribute enhancement, all showing significant performance improvements. For geometry enhancement, our method achieved BD-PSNRs of $9.98$~dB and $7.92$~dB in terms of D1 and D2 metrics, respectively, corresponding to BD-BRs of $-90.98\%$ and $-81.76\%$. For attribute enhancement, the BD-PSNRs for the Y component and YUV were $3.67$~dB and $3.48$~dB, respectively, with corresponding BD-BRs of $-56.88\%$ and $-56.38\%$. In terms of PCQM, our method achieved a gain of $10.83 \times 10^{-3}$ on the BD-PCQM metric and a BD-BR of $-68.35\%$. Additionally, for IWSSIM\textsubscript{p}, we obtained a BD-IWSSIM\textsubscript{p} gain of $17.94 \times 10^{-2}$ and a BD-BR of $-79.26\%$, further validating the superiority of the enhanced point clouds in terms of perceptual quality.

\begin{table}[t]
\centering
\caption{BD-BR and BD-PCQM Gains of UGAE and G-PCC++ Compared to G-PCC}
\label{tab:comparison_with_gpp}
\resizebox{\linewidth}{!}{
\begin{tabular}{lcccc}
\toprule
\multirow{2}{*}{Point Cloud} & \multicolumn{2}{c}{G-PCC++} & \multicolumn{2}{c}{UGAE} \\
\cmidrule(lr){2-3} \cmidrule(lr){4-5}
& BD-BR (\%) & BD-PCQM & BD-BR (\%) & BD-PCQM \\
\midrule
longdress & -38.00 & 0.0028 & -65.39 & 0.0107 \\
loot & -30.91 & 0.0022 & -69.71 & 0.0106 \\
redandblack & -37.27 & 0.0026 & -67.64 & 0.0104 \\
soldier & -33.25 & 0.0027 & -67.43 & 0.0135 \\
\midrule
\textbf{Average} & \textbf{-34.86} & \textbf{0.0026} & \textbf{-67.54} & \textbf{0.0113} \\
\midrule
basketball & -30.99 & 0.0021 & -73.13 & 0.0082 \\
dancer & -30.42 & 0.0019 & -70.24 & 0.0080 \\
exercise & -28.78 & 0.0017 & -69.97 & 0.0067 \\
model & -35.32 & 0.0018 & -74.76 & 0.0097 \\
\midrule
\textbf{Average} & \textbf{-31.38} & \textbf{0.0019} & \textbf{-72.03} & \textbf{0.0082} \\
\midrule
andrew & -26.68 & 0.0018 & -67.13 & 0.0090 \\
david & -20.15 & 0.0010 & -60.63 & 0.0071 \\
phil & -28.38 & 0.0021 & -76.40 & 0.0140 \\
ricardo & -24.41 & 0.0010 & -44.39 & 0.0041 \\
sarah & -29.01 & 0.0014 & -67.57 & 0.0066 \\
\midrule
\textbf{Average} & \textbf{-25.73} & \textbf{0.0015} & \textbf{-63.22} & \textbf{0.0082} \\
\bottomrule
\end{tabular}
}
\end{table}

\begin{table}[t]
\centering
\caption{Processing Time for Different Point Clouds}
\label{tab:runtime}
\begin{tabular}{lccc}
\toprule
{Point Cloud} & {PoGE$^*$ (s)} & {PAE (s)} & {PoAE (s)} \\
\midrule
longdress & 31.25 & 15.30 & 0.19 \\
loot & 30.16 & 13.67 & 0.18 \\
redandblack & 28.18 & 15.08 & 0.17 \\
soldier & 40.94 & 17.76 & 0.23 \\
\midrule
\textbf{Average} & \textbf{32.63} & \textbf{15.45} & \textbf{0.19} \\
\midrule
basketball & 121.90 & 45.15 & 1.71 \\
dancer  & 123.26 & 46.02 & 1.51 \\
exercise  & 120.57 & 40.06 & 1.40 \\
model  & 116.47 & 39.67 & 1.42 \\
\midrule
\textbf{Average}  & \textbf{120.10} & \textbf{42.73} & \textbf{1.51} \\
\midrule
andrew    & 48.39 & 20.82 & 0.27 \\
david     & 55.80 & 23.62 & 0.43 \\
phil      & 60.52 & 25.95 & 0.48 \\
ricardo   & 33.81 & 15.15 & 0.20 \\
sarah     & 50.61 & 26.83 & 0.28 \\
\midrule
\textbf{Average} & \textbf{49.83} & \textbf{22.47} & \textbf{0.33} \\
\bottomrule
\end{tabular}
\\[0.5em]
\scriptsize{*To ensure reproducibility, PoGE was run on CPU, resulting in longer runtime.}
\end{table}

Fig.~\ref{fig:rd_curves} illustrates the overall R-D curves for all tested point clouds. UGAE consistently outperformed the baseline method at different bitrates, especially at low bitrates. This was mainly due to the use of enhanced geometry instead of lossy geometry in the attribute recoloring, which preserved more attribute details while maintaining the same geometry bitrate.

As shown in Table~\ref{tab:comparison_with_gpp}, we conducted a comprehensive comparison between UGAE and the current state-of-the-art geometry-attributes joint enhancement method, G-PCC++. While G-PCC++ aims to post-process both lossy geometry and lossy attributes after reconstruction, it achieved only marginal improvements on the BD-PCQM metric across the three test datasets, with gains of merely $2.6 \times 10^{-3}$, $1.9 \times 10^{-3}$, and $1.5 \times 10^{-3}$, corresponding to BD-BRs of $-34.86\%$, $-31.38\%$, and $-25.73\%$, respectively. In contrast, the average BD-PCQM of UGAE was more than four times that of G-PCC++, corresponding to BD-BRs of $-67.54\%$, $-72.03\%$, and $-63.22\%$,respectively, clearly demonstrating its significant advantages in enhancing both geometry and attribute quality.

\begin{figure*}[t]
    \centering
    \includegraphics[width=0.85\linewidth]{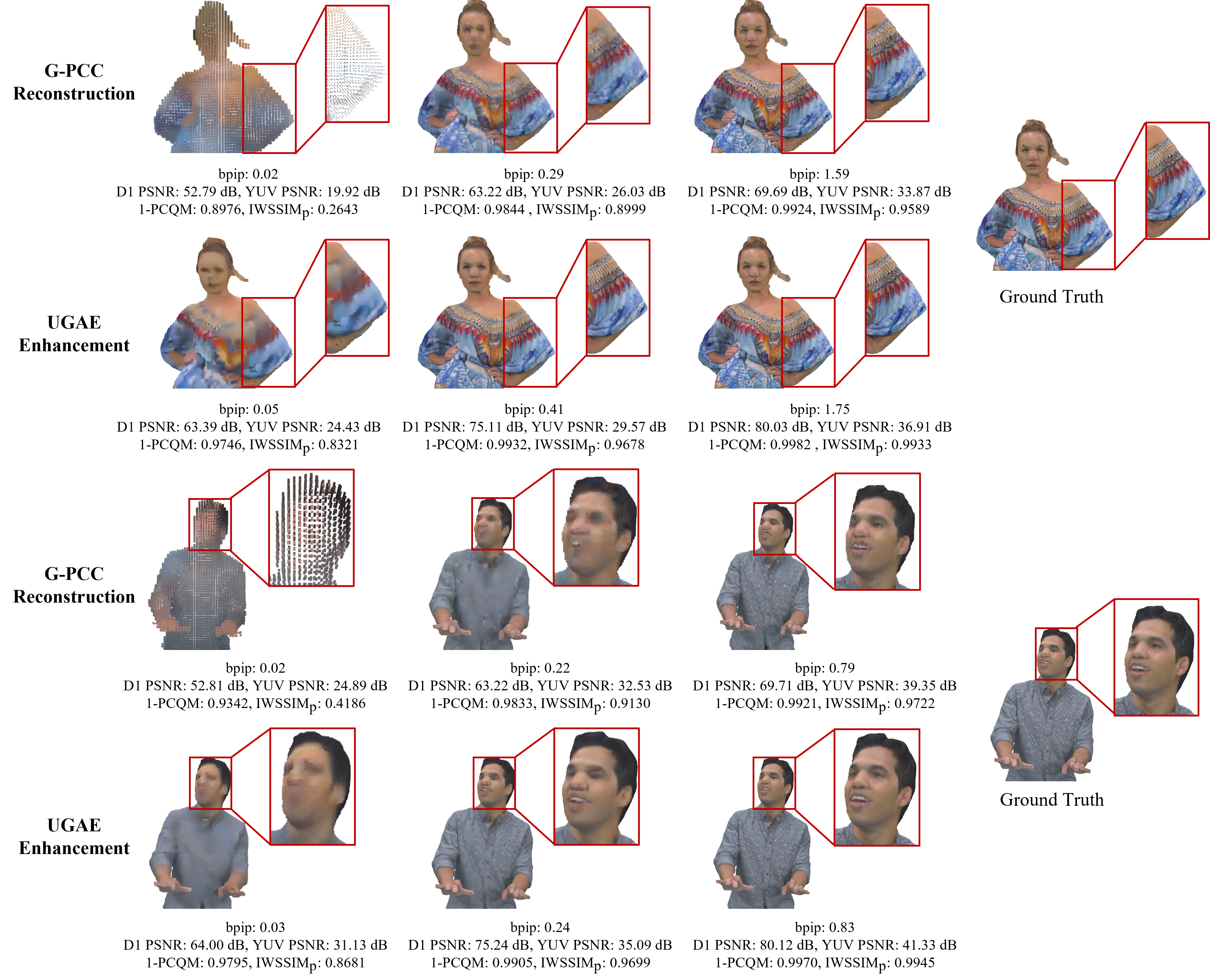}
    \caption{Subjective quality comparison. We selected three bitrates, R01, R03, and R05, to compare the reconstructed point clouds of G-PCC and UGAE.}
    \label{fig:visual_comparison}
\end{figure*}

\begin{table}[t]
\centering
\caption{Geometry Enhancement Performance of PoGE against DGPP.}
\label{tab:po_ge_comparison}
\begin{tabular}{lcccc}
\toprule
\multirow{3}{*}{Point Cloud} & \multicolumn{2}{c}{D1} & \multicolumn{2}{c}{D2} \\
\cmidrule(lr){2-3} \cmidrule(lr){4-5}
 & BD-BR  & BD-PSNR  & BD-BR & BD-PSNR  \\
 & (\%) & (dB) & (\%) & (dB) \\
\midrule
longdress & -28.08 & 1.47 & -47.93 & 2.74 \\
loot & -3.45 & -0.23 & -1.10 & -0.76 \\
redandblack & -27.86 & 1.40 & -30.45 & 1.69 \\
soldier & -27.33 & 1.41 & -29.07 & 1.72 \\
\midrule
\textbf{Average} & \textbf{-21.68} & \textbf{1.01} & \textbf{-27.14} & \textbf{1.35} \\
\midrule
basketball & -26.87 & 1.24 & -19.01 & 1.07 \\
dancer & -28.09 & 1.36 & -22.61 & 1.29 \\
exercise & -26.22 & 1.30 & -22.87 & 1.34 \\
model & -27.00 & 1.33 & -22.75 & 1.24 \\
\midrule
\textbf{Average} & \textbf{-27.04} & \textbf{1.31} & \textbf{-21.81} & \textbf{1.24} \\
\midrule
andrew & -45.06 & 2.01 & -23.10 & 0.97 \\
david & -45.49 & 2.01 & -28.05 & 1.27 \\
phil & -40.86 & 1.80 & -21.12 & 0.99 \\
ricardo & -46.90 & 2.25 & -22.00 & 0.97 \\
sarah & -44.42 & 2.10 & -25.09 & 1.24 \\
\midrule
\textbf{Average} & \textbf{-44.55} & \textbf{2.04} & \textbf{-23.87} & \textbf{1.09} \\
\bottomrule
\end{tabular}
\end{table}

\begin{table*}
\centering
\caption{Geometry Enhancement Gains: Dense Connection vs. IRN Against G-PCC.}
\label{tab:dense_connection_ablation}
\resizebox{0.8\linewidth}{!}{
\scriptsize
\begin{tabular}{lcccccccc}
\toprule
\multirow{4}{*}{Point Cloud}
& \multicolumn{4}{c}{IRN}
& \multicolumn{4}{c}{Dense Connection} \\
\cmidrule(r){2-5} \cmidrule(l){6-9}
& \multicolumn{2}{c}{D1} & \multicolumn{2}{c}{D2} & \multicolumn{2}{c}{D1} & \multicolumn{2}{c}{D2} \\
\cmidrule(r){2-3} \cmidrule(r){4-5} \cmidrule(r){6-7} \cmidrule(r){8-9} 
& BD-BR & BD-PSNR & BD-BR & BD-PSNR & BD-BR & BD-PSNR & BD-BR & BD-PSNR  \\
& (\%) & (dB) & (\%) & (dB) & (\%) & (dB) & (\%) & (dB) \\
\midrule
longdress     & -84.07 & 8.56 & -75.19 & 6.85 & \textbf{-88.71} & \textbf{9.44} & \textbf{-77.14} & \textbf{7.33} \\
loot          & -86.36 & 9.67 & -79.86 & 8.27 & \textbf{-90.43} & \textbf{10.72} & \textbf{-82.32} & \textbf{8.81} \\
redandblack   & -83.10 & 8.67 & -76.48 & 7.32 & \textbf{-88.23} & \textbf{9.55} & \textbf{-79.00} & \textbf{7.73} \\
soldier       & -85.73 & 9.24 & -77.27 & 7.70 & \textbf{-88.58} & \textbf{9.91} & \textbf{-78.56} & \textbf{8.03} \\
\midrule
\textbf{Average} & -84.82 & 9.04 & -77.20 & 7.43 & \textbf{-88.99} & \textbf{9.91} & \textbf{-79.26} & \textbf{7.98} \\
\bottomrule
\end{tabular}
}
\end{table*}

\subsection{Subjective Quality Evaluation}
Fig.~\ref{fig:visual_comparison} presents a visual comparison of the G-PCC reconstructed point clouds, the UGAE-enhanced point clouds, and the original point clouds at different bitrates. At the low bitrate, the G-PCC reconstructed point clouds exhibited severe geometry and color distortions, making the overall object outlines difficult to recognize. In contrast, the UGAE-enhanced point clouds showed geometry structures much closer to the original ones. Although the texture information was blurred, the enhanced results allowed for the distinction of attribute variations across different regions. At the medium bitrate, the reconstructed point clouds of G-PCC suffered from surface roughness and loss of high-frequency details. However, the UGAE-enhanced results demonstrates smoother geometry surfaces and richer texture details. At the high bitrate, the geometry of the reconstructed point clouds from G-PCC remained rough, and the texture details in high-frequency regions were still difficult to discern. In contrast, the UGAE-enhanced point clouds closely approximated the original point clouds in both geometry and attributes, demonstrating superior reconstruction quality. More visual results are provided in the supplementary materials. 

\subsection{Time Complexity}
Table~\ref{tab:runtime} provides the average run times of each component of UGAE across different datasets. To ensure reproducibility of geometry enhancement, we ran the TSConv layer of PoGE on the CPU, resulting in relatively longer execution times for this part. For PAE, only the time reported corresponds to the DA-KNN recoloring, as the time for PoGE has already been provided separately. Since Owlii contains the most points, its recoloring time was the highest. PoAE was entirely executed on the GPU and processed the entire point cloud in a single forward pass, leading to high computational efficiency. On the three datasets, the average processing times of PoAE were $0.19$ s (8iVFB), $1.51$ s (Owlii), and $0.33$ s (MVUB).

\subsection{Ablation Study}
In this section, we study the effectiveness of the core elements of UGAE: PoGE, PAE, PoAE, and the proposed W-MSE loss function.

\begin{table*}[t]
\centering
\caption{Attribute Enhancement Gains of PAE, PoAE, and W-MSE Over G-PCC.}
\label{tab:pae_only}
\begin{tabular}{lcccccccccccc}
\toprule
\multirow{5}{*}{Point Cloud} & \multicolumn{4}{c}{PAE} & \multicolumn{4}{c}{PAE+PoAE+MSE} & \multicolumn{4}{c}{PAE+PoAE+WMSE} \\
\cmidrule(r){2-5} \cmidrule(r){6-9} \cmidrule(r){10-13}
& \multicolumn{2}{c}{Y} & \multicolumn{2}{c}{YUV} & \multicolumn{2}{c}{Y} & \multicolumn{2}{c}{YUV} & \multicolumn{2}{c}{Y} & \multicolumn{2}{c}{YUV} \\
\cmidrule(r){2-3} \cmidrule(r){4-5} \cmidrule(r){6-7} \cmidrule(r){8-9} \cmidrule(r){10-11} \cmidrule(r){12-13}
& BD-  & BD- & BD- & BD- & BD-  & BD- & BD- & BD- & BD-  & BD- & BD- & BD-  \\
&BR &PSNR &BR &PSNR &BR &PSNR &BR &PSNR &BR &PSNR &BR &PSNR \\
& (\%) & (dB) &(\%) & (dB) & (\%) & (dB) &(\%) & (dB) & (\%) & (dB) &(\%) & (dB) \\
\midrule
longdress    & -39.63 & 1.88 & -38.33 & 1.82 & -48.89 & 2.44 & -47.90 & 2.38 & \textbf{-49.74} & \textbf{2.49} & \textbf{-49.00} & \textbf{2.45} \\
loot         & -40.32 & 2.40 & -39.60 & 2.34 & -47.38 & 2.91 & -47.89 & 2.90 & \textbf{-49.50} & \textbf{2.99} & \textbf{-50.31} & \textbf{3.00} \\
redandblack  & -41.27 & 2.25 & -40.28 & 2.18 & -51.33 & 2.92 & -50.24 & 2.83 & \textbf{-52.74} & \textbf{2.99} & \textbf{-51.49} & \textbf{2.89} \\
soldier      & -37.12 & 2.15 & -35.82 & 2.00 & -46.86 & 2.79 & -46.08 & 2.63 & \textbf{-47.95} & \textbf{2.86} & \textbf{-47.27} & \textbf{2.69} \\
\midrule
\textbf{Average} & -39.58 & 2.17 & -38.51 & 2.08 & -48.62 & 2.77 & -48.03 & 2.69 & \textbf{-49.99} & \textbf{2.83} & \textbf{-49.52} & \textbf{2.76} \\
\bottomrule
\end{tabular}
\end{table*}

\begin{table*}
\centering
\caption{Sensitivity Analysis of DA-KNN Parameter $k$ on PAE Recoloring Performance.}
\label{tab:knn_ablation}
\begin{tabular}{lcccccccccccc}
\toprule
\multirow{5}{*}{Point Cloud} & \multicolumn{4}{c}{$k=4$} & \multicolumn{4}{c}{$k=8$} & \multicolumn{4}{c}{$k=16$} \\
\cmidrule(r){2-5} \cmidrule(r){6-9} \cmidrule(r){10-13}
& \multicolumn{2}{c}{Y} & \multicolumn{2}{c}{YUV} & \multicolumn{2}{c}{Y} & \multicolumn{2}{c}{YUV} & \multicolumn{2}{c}{Y} & \multicolumn{2}{c}{YUV} \\
\cmidrule(r){2-3} \cmidrule(r){4-5} \cmidrule(r){6-7} \cmidrule(r){8-9} \cmidrule(r){10-11} \cmidrule(r){12-13}
& BD-  & BD- & BD- & BD- & BD-  & BD- & BD- & BD- & BD-  & BD- & BD- & BD-  \\
&BR &PSNR &BR &PSNR &BR &PSNR &BR &PSNR &BR &PSNR &BR &PSNR \\
& (\%) & (dB) &(\%) & (dB) & (\%) & (dB) &(\%) & (dB) & (\%) & (dB) &(\%) & (dB) \\
\midrule
longdress     & -49.74 & 2.49 & -48.99 & 2.45 & \textbf{-49.74} & \textbf{2.49} & \textbf{-49.00} & \textbf{2.45} & -49.75 & 2.49 & -48.99 & 2.45 \\
loot          & -48.83 & 2.94 & -49.60 & 2.95 & \textbf{-49.50} & \textbf{2.99} & \textbf{-50.31} & \textbf{3.00} & -48.85 & 2.94 & -49.61 & 2.95 \\
redandblack   & -52.07 & 2.94 & -50.79 & 2.84 & \textbf{-52.74} & \textbf{2.99} & \textbf{-51.49} & \textbf{2.89} & -52.06 & 2.94 & -50.78 & 2.84 \\
soldier       & -47.36 & 2.80 & -46.65 & 2.64 & \textbf{-47.95} & \textbf{2.86} & \textbf{-47.27} & \textbf{2.69} & -47.43 & 2.81 & -46.69 & 2.65 \\
\midrule
\textbf{Average} & -49.50 & 2.79 & -49.00 & 2.72 & \textbf{-49.99} & \textbf{2.83} & \textbf{-49.52} & \textbf{2.76} & -50.02 & 2.83 & -49.54 & 2.76 \\
\bottomrule
\end{tabular}
\end{table*}

First, we compare the proposed PoGE with the post-geometry enhancement network DGPP. As shown in Table~\ref{tab:po_ge_comparison}, PoGE achieved performance gains of $1.01$~dB (resp. $1.35$~dB), $1.31$~dB (resp. $1.24$~dB), and $2.04$~dB (resp. $1.09$~dB) in terms of BD-PSNR of the D1 (resp. D2) metric over DGPP \cite{ref58} on the three test datasets. In terms of BD-BR, PoGE achieved bitrate reductions of $-21.68\%$ (resp. $-27.14\%$), $27.04\%$ (resp. $-21.81\%$), and $-44.55\%$ (resp. $-23.87\%$), respectively, on the three test datasets.

As shown in Fig.~\ref{fig:poge}, after TSConv upsamples the extracted features, it is necessary to reduce the high-dimensional features to a single dimension to predict occupancy probability. To retain more geometry information, some works~\cite{ref41, ref47} introduces InceptionResNet (IRN) to enhance the features. However, we used dense connection to fuse multi-dimensional features. To validate the effectiveness of this approach, we compared PoGE using IRN and PoGE using dense connections against the lossy geometry reconstructed by G-PCC. The results, shown in Table~\ref{tab:dense_connection_ablation}, indicate that on the 8iVFB dataset, dense connections achieved an additional BD-BRs of $-4.17\%$ (D1) and $-2.06\%$ (D2), along with BD-PSNR gains of $0.87$~dB and $0.37$~dB, compared to IRN.

To evaluate the effectiveness of recoloring based on the enhanced geometry, we removed PoAE and retained only PAE for attribute reconstruction. As shown in the left part of Table~\ref{tab:pae_only}, UGAE without PoAE still achieved a BD-PSNR gain of $2.17$~dB on the Y component and $2.08$~dB on YUV, demonstrating that PAE can effectively improve the attribute quality by leveraging the enhanced geometry. Furthermore, we analyzed how the number of neighbors $k$ in the recoloring process affects overall performance. As shown in Table~\ref{tab:knn_ablation}, with $k = 4, 8, 16$, PoAE achieved nearly identical BD-PSNR at $k=8$ and $k=16$. The corresponding average run times were $14.23$ s ($k=4$), $15.45$ s ($k=8$), and $19.23$ s ($k=16$). Considering the trade-off between quality improvement and time consumption, we selected $k=8$ as the default setting in DA-KNN.

Building upon PAE, we further introduce PoAE to enhance reconstructed attributes. As shown in the right part of Table~\ref{tab:pae_only}, after adding PoAE, the BD-PSNR was further improved by $0.6$~dB on Y and $0.61$~dB on the YUV components, validating the positive contribution of PoAE in enhancing the quality of the reconstructed attributes. 

Furthermore, to evaluate the effectiveness of the proposed W-MSE loss function, we replaced it with the standard, unweighted MSE. Comparing the middle and right parts of Table~\ref{tab:pae_only}, W-MSE achieved an approximate BD-PSNR gain of $0.07$~dB over MSE. It is worth noting that, although the standard MSE already imposes some penalty on regions with large distortion, W-MSE still demonstrates a strong complementary enhancement effect.

\section{Conclusion}
\label{sec:conclusion}
We proposed UGAE, a joint enhancement framework for point cloud compression under both lossy geometry and lossy attribute conditions. The framework consists of three main components: PoGE, PAE, and PoAE. PoGE combines Transformer and U-Net architectures to effectively enhance geometry. PAE improves attribute compression quality by recoloring based on the original attributes and the enhanced geometry at the encoder side. PoAE further refines the reconstructed attributes at the decoder side by using the proposed W-MSE loss function, which compensates for the loss in high-frequency details. Extensive experimental results demonstrate that UGAE outperforms state-of-the-art methods, achieving substantial improvements across various geometry and attribute quality metrics. Future work will focus on further optimizing computational efficiency, exploring more effective strategies for joint geometry-attribute enhancement, and extending UGAE to broader application scenarios, such as LiDAR point clouds and dynamic point clouds.

\begin{IEEEbiography}[{\includegraphics[width=1in,height=1.25in,clip,keepaspectratio]{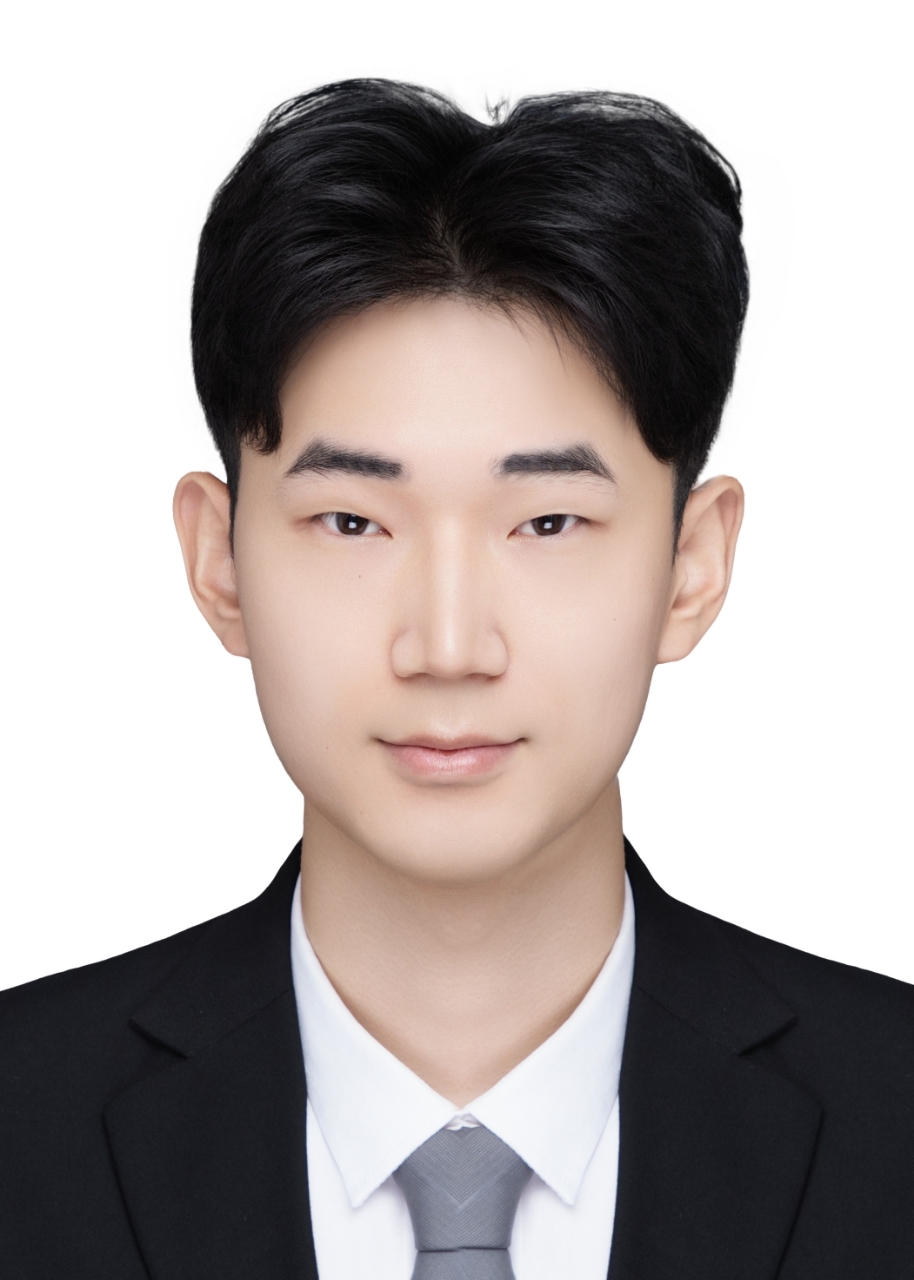}}]
{Pan Zhao} 
received the B.E. degree from the School of Software Engineering, Jinling Institute of Technology, Nanjing, China, in 2021, and the M.S. degree from School of Artificial Intelligence, Nanjing University of Information Science and Technology, Nanjing, China, in 2024. He is currently pursuing the Ph.D. degree with the School of Control Science and Engineering, Shandong University, Jinan, China. His research interests include point cloud compression and quality enhancement.
\end{IEEEbiography}

\vspace{-0.5cm}
\begin{IEEEbiography}[{\includegraphics[width=1in,height=1.25in,clip,keepaspectratio]{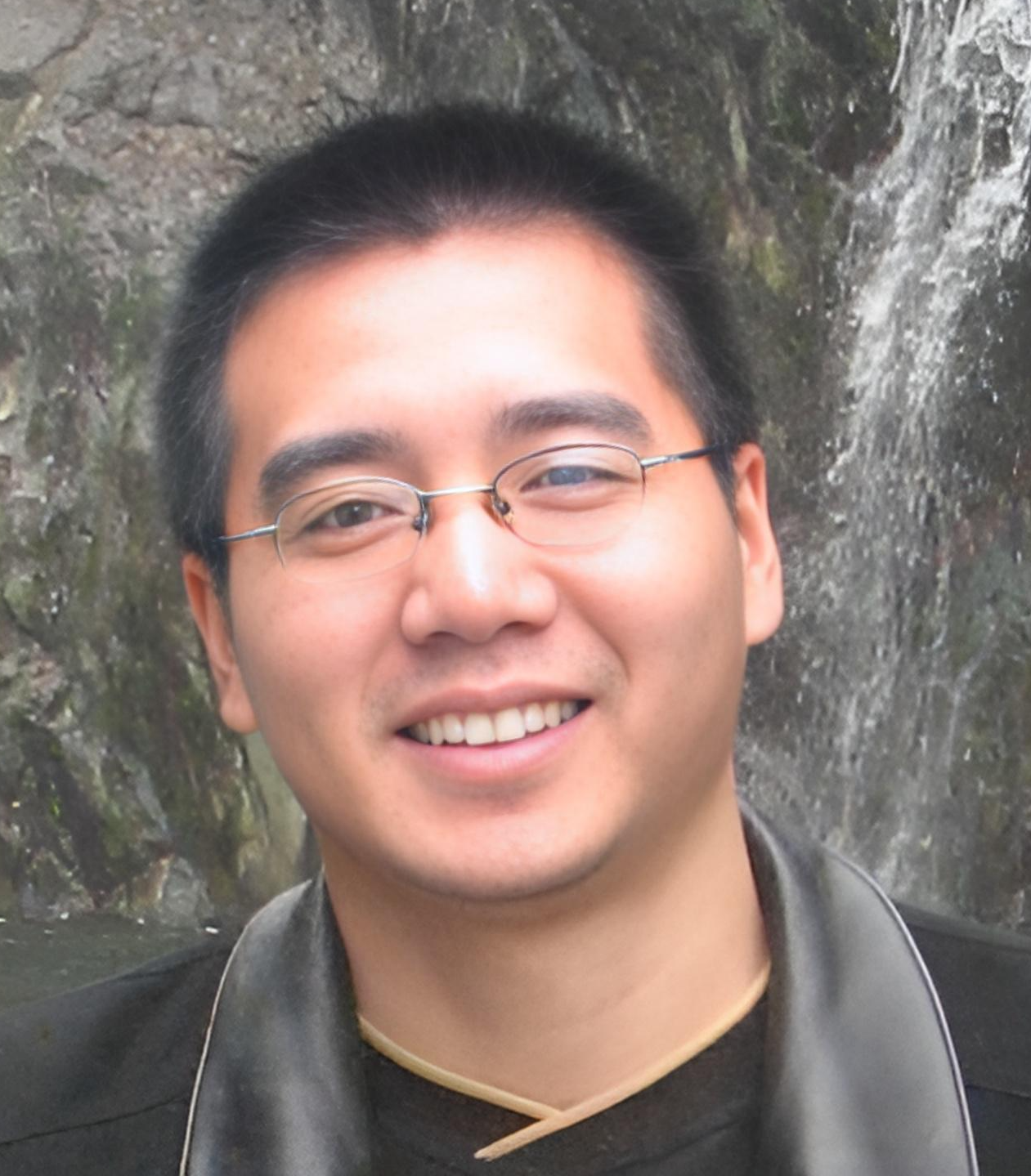}}]
{Hui Yuan} (Senior Member, IEEE) received the B.E. and Ph.D. degrees in telecommunication engineering from Xidian University, Xi’an, China, in 2006 and 2011, respectively. In April 2011, he joined Shandong University, Ji’nan, China, as a Lecturer (April 2011–December 2014), an Associate Professor (January 2015-October 2016), and a Professor (September 2016). From January 2013-December 2014, and November 2017-February 2018, he also worked as a Postdoctoral Fellow (Granted by the Hong Kong Scholar Project) and a Research Fellow, respectively, with the Department of Computer Science, City University of Hong Kong, Hong Kong. From November 2020 to November 2021, he also worked as a Marie Curie Fellow (Granted by the Marie Skłodowska-Curie Individual Fellowships of European Commission) with the Faculty of Computing, Engineering and Media, De Montfort University, United Kingdom. From October 2021 to November 2021, he also worked as a visiting researcher (secondment of the Marie Skłodowska-Curie Individual Fellowships) with the Computer Vision and Graphics group, Fraunhofer Heinrich-Hertz-Institut (HHI), Germany. His current research interests include 3D visual coding, processing, and communication. He is serving as an Associate Editor for IEEE Transactions on Image Processing, IEEE Transactions on Consumer Electronics, and IET Image Processing, an Area Chair for IEEE ICME.
\end{IEEEbiography}

\vspace{-0.5cm}
\begin{IEEEbiography}[{\includegraphics[width=1in,height=1.25in,clip,keepaspectratio]{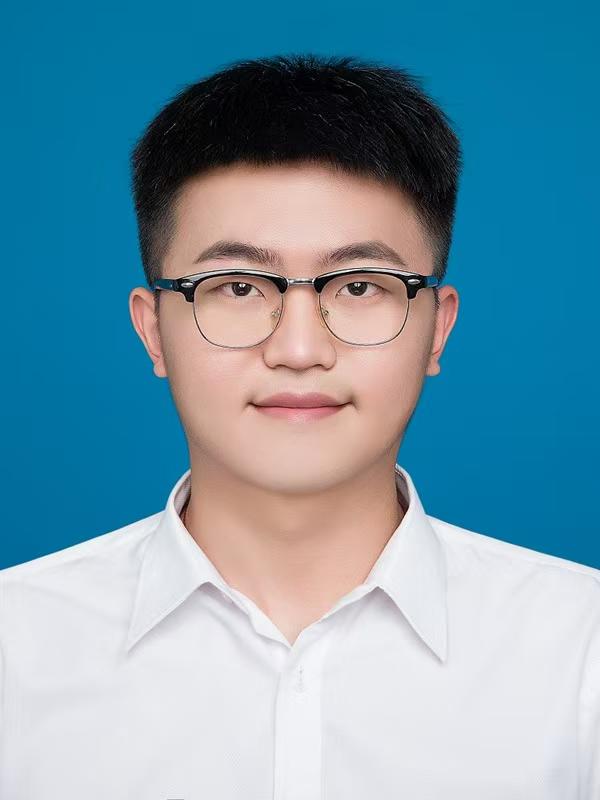}}]
{Chongzhen Tian} received the B.S. degree and M.S. degree from Ningbo University, Ningbo, China, in 2020 and 2023, respectively. He is currently pursuing the Ph.D. degree at the Shandong University, Jinan, China. His research interests include point cloud compression and quality assessment.	
\end{IEEEbiography}

\IEEEaftertitletext{\vspace{-2\baselineskip}}

\vspace{-0.5cm}
\begin{IEEEbiography}[{\includegraphics[width=1in,height=1.25in,clip,keepaspectratio]{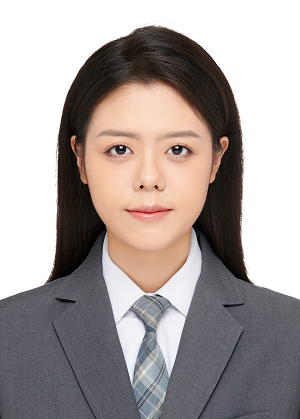}}]
{Tian Guo}
 received the B.E. degree from the School of Information and Control Engineering, China University of Mining and Technology, Xuzhou, China, in 2021. She is currently pursuing the Ph.D. degree with the School of Control Science and Engineering, Shandong University, Jinan, China. Her research interests include point cloud compression and processing.		
\end{IEEEbiography}

\vspace{-0.5cm}
\begin{IEEEbiography}[{\includegraphics[width=1in,height=1.25in,clip,keepaspectratio]{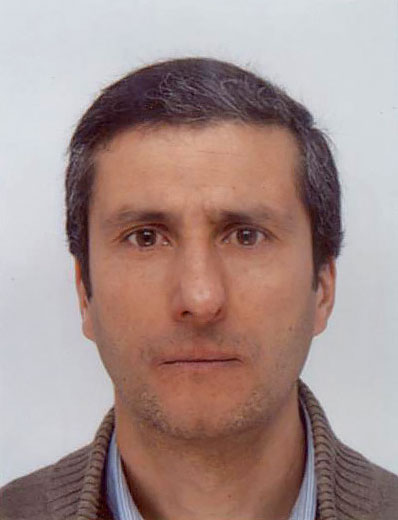}}]
{Raouf Hamzaoui} (Senior Member, IEEE) received
the M.Sc. degree in mathematics from the University of Montreal, Canada, in 1993, and the Dr.rer.nat.degree from the University of Freiburg, Germany, in 1997, and the Habilitation degree in computer science from the University of Konstanz, Germany, in 2004. He was an Assistant Professor with the Department of Computer Science, University of Leipzig, Germany, and the Department of Computer and Information Science, University of Konstanz. In September 2006, he joined De Montfort University, where he is currently a Professor in media technology. He was a member of the Editorial Board of the IEEE TRANSACTIONS ON MULTIMEDIA and IEEE TRANSACTIONS ON CIRCUITS AND SYSTEMS FOR VIDEO TECHNOLOGY. He has published more than 120 research papers in books, journals, and conferences. His research has been funded by the EU, DFG, Royal Society, and industry and received best paper awards (ICME 2002, PV’07, CONTENT 2010, MESM’2012, and UIC-2019).
\end{IEEEbiography}

\vspace{-0.5cm}
\begin{IEEEbiography}[{\includegraphics[width=1in,height=1.25in,clip,keepaspectratio]{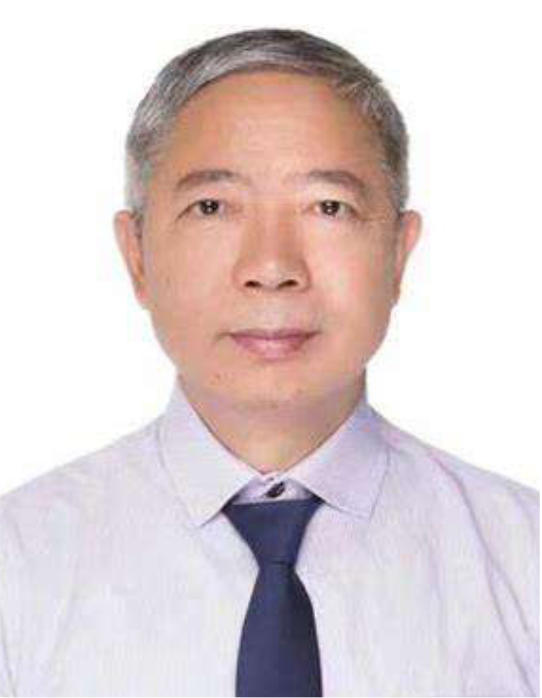}}]
{Zhigeng Pan} received the Ph.D. degree in computer graphics from Zhejiang University, Hangzhou, China, in 1993. He is currently the Dean of the School of Artificial Intelligence (School of Future Technology), Nanjing University of Information Science and Technology, Nanjing, China. His research interests include virtual reality, computer graphics, and human-computer interaction.
\end{IEEEbiography}


\begin{thebibliography}{99}
\bibliographystyle{IEEEtran}


\bibitem{ref1}
B. Mildenhall, P. P. Srinivasan, M. Tancik, J. T. Barron, R. Ramamoorthi, and R. Ng, “Nerf: Representing scenes as neural radiance fields for view synthesis,” \emph{Commun ACM}, vol. 65, no. 1, pp. 99–106, 2021.

\bibitem{ref2}
S. Schwarz et al., “Emerging MPEG standards for point cloud compression,” \emph{IEEE J Emerg Sel Top Circuits Syst}, vol. 9, no. 1, pp. 133–148, 2018.

\bibitem{ref3}
C. Cao, M. Preda, V. Zakharchenko, E. S. Jang, and T. Zaharia, “Compression of sparse and dense dynamic point clouds—methods and standards,” \emph{Proceedings of the IEEE}, vol. 109, no. 9, pp. 1537–1558, 2021.

\bibitem{ref4}
D. Graziosi, O. Nakagami, S. Kuma, A. Zaghetto, T. Suzuki, and A. Tabatabai, “An overview of ongoing point cloud compression standardization activities: Video-based (V-PCC) and geometry-based (G-PCC),” \emph{APSIPA Trans Signal Inf Process}, vol. 9, p. e13, 2020.

\bibitem{ref5}
K. Mammou, P. A. Chou, D. Flynn, M. Krivoku\'{c}a, O. Nakagami, and T. Sugio, 
“G-PCC codec description v2,” document ISO/IEC JTC1/SC29/WG11 N18189, 2019.

\bibitem{ref6}
K. Mammou, A. M. Tourapis, D. Singer, and Y. Su, 
“Video-based and hierarchical approaches point cloud compression,” 
document ISO/IEC JTC1/SC29/WG11 m41649, Macau, China, 2017.

\bibitem{ref7}
L. Gao, T. Fan, J. Wan, Y. Xu, J. Sun, and Z. Ma, “Point cloud geometry compression via neural graph sampling,” in \emph{2021 IEEE International Conference on Image Processing (ICIP)}, IEEE, 2021, pp. 3373–3377.

\bibitem{ref8}
T. Huang and Y. Liu, “3d point cloud geometry compression on deep learning,” in \emph{Proceedings of the 27th ACM international conference on multimedia}, 2019, pp. 890–898.

\bibitem{ref9}
X. Wen, X. Wang, J. Hou, L. Ma, Y. Zhou, and J. Jiang, “Lossy geometry compression of 3d point cloud data via an adaptive octree-guided network,” in \emph{2020 IEEE International Conference on Multimedia and Expo (ICME)}, IEEE, 2020, pp. 1–6.

\bibitem{ref10}
W. Yan, S. Liu, T. H. Li, Z. Li, and G. Li, “Deep autoencoder-based lossy geometry compression for point clouds,” arXiv preprint arXiv:1905.03691, 2019.

\bibitem{ref11}
K. You and P. Gao, “Patch-based deep autoencoder for point cloud geometry compression,” in \emph{Proceedings of the 3rd ACM International Conference on Multimedia in Asia}, 2021, pp. 1–7.

\bibitem{ref12}
K. You, P. Gao, and Q. Li, “IPDAE: Improved patch-based deep autoencoder for lossy point cloud geometry compression,” in \emph{Proceedings of the 1st International Workshop on Advances in Point Cloud Compression, Processing and Analysis}, 2022, pp. 1–10.

\bibitem{ref13}
A. F. R. Guarda, N. M. M. Rodrigues, and F. Pereira, “Adaptive deep learning-based point cloud geometry coding,” \emph{IEEE J Sel Top Signal Process}, vol. 15, no. 2, pp. 415–430, 2020.

\bibitem{ref14}
M. Quach, G. Valenzise, and F. Dufaux, “Learning convolutional transforms for lossy point cloud geometry compression,” in \emph{2019 IEEE international conference on image processing (ICIP)}, IEEE, 2019, pp. 4320–4324.

\bibitem{ref15}
J. Wang, H. Zhu, H. Liu, and Z. Ma, “Lossy point cloud geometry compression via end-to-end learning,” \emph{IEEE Transactions on Circuits and Systems for Video Technology}, vol. 31, no. 12, pp. 4909–4923, 2021.

\bibitem{ref16}
C. Sun, H. Yuan, S. Li, X. Lu, and R. Hamzaoui, “Enhancing Context Models for Point Cloud Geometry Compression With Context Feature Residuals and Multi-Loss,” \emph{IEEE J Emerg Sel Top Circuits Syst}, vol. 14, no. 2, pp. 224–234, 2024.

\bibitem{ref17}
L. Huang, S. Wang, K. Wong, J. Liu, and R. Urtasun, “Octsqueeze: Octree-structured entropy model for lidar compression,” in \emph{Proceedings of the IEEE/CVF conference on computer vision and pattern recognition}, 2020, pp. 1313–1323.

\bibitem{ref18}
S. Biswas, J. Liu, K. Wong, S. Wang, and R. Urtasun, “Muscle: Multi sweep compression of lidar using deep entropy models,” \emph{Adv Neural Inf Process Syst}, vol. 33, pp. 22170–22181, 2020.

\bibitem{ref19}
Z. Que, G. Lu, and D. Xu, “Voxelcontext-net: An octree based framework for point cloud compression,” in \emph{Proceedings of the IEEE/CVF Conference on Computer Vision and Pattern Recognition}, 2021, pp. 6042–6051.

\bibitem{ref20}
Z. Chen, Z. Qian, S. Wang, and Q. Chen, “Point cloud compression with sibling context and surface priors,” in \emph{European Conference on Computer Vision}, Springer, 2022, pp. 744–759.

\bibitem{ref21}
C. Fu, G. Li, R. Song, W. Gao, and S. Liu, “Octattention: Octree-based large-scale contexts model for point cloud compression,” in \emph{Proceedings of the AAAI conference on artificial intelligence}, 2022, pp. 625–633.

\bibitem{ref22}
R. Song, C. Fu, S. Liu, and G. Li, “Efficient hierarchical entropy model for learned point cloud compression,” in \emph{Proceedings of the IEEE/CVF Conference on Computer Vision and Pattern Recognition}, 2023, pp. 14368–14377.

\bibitem{ref23}
M. Cui, J. Long, M. Feng, B. Li, and H. Kai, “OctFormer: Efficient octree-based transformer for point cloud compression with local enhancement,” in \emph{Proceedings of the AAAI Conference on Artificial Intelligence}, 2023, pp. 470–478.

\bibitem{ref24}
C. Sun, H. Yuan, X. Mao, X. Lu, and R. Hamzaoui, “Enhancing Octree-Based Context Models for Point Cloud Geometry Compression With Attention-Based Child Node Number Prediction,” \emph{IEEE Signal Process Lett}, vol. 31, pp. 1835–1839, 2024.

\bibitem{ref25}
J. Wang, D. Ding, Z. Li, and Z. Ma, “Multiscale point cloud geometry compression,” in \emph{2021 Data Compression Conference (DCC)}, IEEE, 2021, pp. 73–82.

\bibitem{ref26}
S. Xia, T. Fan, Y. Xu, J.-N. Hwang, and Z. Li, “Learning dynamic point cloud compression via hierarchical inter-frame block matching,” in \emph{Proceedings of the 31st ACM International Conference on Multimedia}, 2023, pp. 7993–8003.

\bibitem{ref27}
J. Wang, D. Ding, Z. Li, X. Feng, C. Cao, and Z. Ma, “Sparse tensor-based multiscale representation for point cloud geometry compression,” \emph{IEEE Trans Pattern Anal Mach Intell}, vol. 45, no. 7, pp. 9055–9071, 2022.

\bibitem{ref28}
J. Wang, R. Xue, J. Li, D. Ding, Y. Lin, and Z. Ma, “A versatile point cloud compressor using universal multiscale conditional coding–Part I: Geometry,” \emph{IEEE Trans Pattern Anal Mach Intell}, 2024.

\bibitem{ref29}
J. Wang, D. Ding, and Z. Ma, “Lossless point cloud attribute compression using cross-scale, cross-group, and cross-color prediction,” in \emph{2023 Data Compression Conference (DCC)}, IEEE, 2023, pp. 228–237.

\bibitem{ref30}
D. T. Nguyen and A. Kaup, “Lossless point cloud geometry and attribute compression using a learned conditional probability model,” \emph{IEEE Transactions on Circuits and Systems for Video Technology}, vol. 33, no. 8, pp. 4337–4348, 2023.

\bibitem{ref31}
D. T. Nguyen, K. G. Nambiar, and A. Kaup, “Deep probabilistic model for lossless scalable point cloud attribute compression,” in \emph{ICASSP 2023 - 2023 IEEE International Conference on Acoustics, Speech and Signal Processing (ICASSP)}, IEEE, 2023, pp. 1–5.

\bibitem{ref32}
J. Wang, R. Xue, J. Li, D. Ding, Y. Lin, and Z. Ma, “A Versatile Point Cloud Compressor Using Universal Multiscale Conditional Coding–Part II: Attribute,” \emph{IEEE Trans Pattern Anal Mach Intell}, vol. 47, no. 1, pp. 252–268, 2025.

\bibitem{ref33}
X. Sheng, L. Li, D. Liu, Z. Xiong, Z. Li, and F. Wu, “Deep-PCAC: An end-to-end deep lossy compression framework for point cloud attributes,” \emph{IEEE Trans Multimedia}, vol. 24, pp. 2617–2632, 2021.

\bibitem{ref34}
J. Wang and Z. Ma, “Sparse tensor-based point cloud attribute compression,” in \emph{2022 IEEE 5th International Conference on Multimedia Information Processing and Retrieval (MIPR)}, IEEE, 2022, pp. 59–64.

\bibitem{ref35}
G. Fang, Q. Hu, H. Wang, Y. Xu, and Y. Guo, “3dac: Learning attribute compression for point clouds,” in \emph{Proceedings of the IEEE/CVF Conference on Computer Vision and Pattern Recognition}, 2022, pp. 14819–14828.

\bibitem{ref36}
X. Mao, H. Yuan, X. Lu, R. Hamzaoui, and W. Gao, 
“PCAC-GAN: A sparse-tensor-based generative adversarial network for 3D point cloud attribute compression,” 
\emph{arXiv preprint arXiv:2407.05677}, 2024.

\bibitem{ref37}
L. Yu, X. Li, C.-W. Fu, D. Cohen-Or, and P.-A. Heng, “Pu-net: Point cloud upsampling network,” in \emph{Proceedings of the IEEE conference on computer vision and pattern recognition}, 2018, pp. 2790–2799.

\bibitem{ref38}
H. Liu, H. Yuan, J. Hou, R. Hamzaoui, and W. Gao, “PUFA-GAN: A frequency-aware generative adversarial network for 3D point cloud upsampling,” \emph{IEEE Transactions on Image Processing}, vol. 31, pp. 7389–7402, 2022.

\bibitem{ref39}
H. Liu, H. Yuan, R. Hamzaoui, Q. Liu, and S. Li, “PU-Mask: 3D Point Cloud Upsampling via an Implicit Virtual Mask,” \emph{IEEE Transactions on Circuits and Systems for Video Technology}, 2024.

\bibitem{ref40}
A. Akhtar, Z. Li, G. Van der Auwera, L. Li, and J. Chen, “Pu-dense: Sparse tensor-based point cloud geometry upsampling,” \emph{IEEE Transactions on Image Processing}, vol. 31, pp. 4133–4148, 2022.

\bibitem{ref41}
G. Liu, R. Xue, J. Li, D. Ding, and Z. Ma, “Grnet: Geometry restoration for g-pcc compressed point clouds using auxiliary density signaling,” \emph{IEEE Trans Vis Comput Graph}, vol. 30, no. 10, pp. 6740–6753, 2023.

\bibitem{ref42}
J. Zhang, T. Chen, D. Ding, and Z. Ma, “G-PCC++: Enhanced geometry-based point cloud compression,” in \emph{Proceedings of the 31st ACM International Conference on Multimedia}, 2023, pp. 1352–1363.

\bibitem{ref43}
C. Choy, J. Gwak, and S. Savarese, “4d spatio-temporal convnets: Minkowski convolutional neural networks,” in \emph{Proceedings of the IEEE/CVF conference on computer vision and pattern recognition}, 2019, pp. 3075–3084.

\bibitem{ref44}
J. Xing, H. Yuan, W. Zhang, T. Guo, and C. Chen, “A small-scale image U-Net-based color quality enhancement for dense point cloud,” \emph{IEEE Transactions on Consumer Electronics}, 2024.

\bibitem{ref45}
T. Guo, H. Yuan, Q. Liu, H. Su, R. Hamzaoui, and S. Kwong, “PCE-GAN: A Generative Adversarial Network for Point Cloud Attribute Quality Enhancement based on Optimal Transport,” arXiv preprint arXiv:2503.00047, 2025.

\bibitem{ref46}
J. Xing, H. Yuan, R. Hamzaoui, H. Liu, and J. Hou, “GQE-Net: A graph-based quality enhancement network for point cloud color attribute,” \emph{IEEE Transactions on Image Processing}, vol. 32, pp. 6303–6317, 2023.

\bibitem{ref47}
O. Ronneberger, P. Fischer, and T. Brox, “U-net: Convolutional networks for biomedical image segmentation,” in \emph{Medical image computing and computer-assisted intervention–MICCAI 2015: 18th international conference, Munich, Germany, October 5-9, 2015, proceedings, part III 18}, Springer, 2015, pp. 234–241.

\bibitem{ref48}
X. Wu et al., “Point transformer v3: Simpler faster stronger,” in \emph{Proceedings of the IEEE/CVF Conference on Computer Vision and Pattern Recognition}, 2024, pp. 4840–4851.

\bibitem{ref49}
G. Huang, Z. Liu, L. Van Der Maaten, and K. Q. Weinberger, “Densely connected convolutional networks,” in \emph{Proceedings of the IEEE conference on computer vision and pattern recognition}, 2017, pp. 4700–4708.

\bibitem{ref50}
R. L. De Queiroz and P. A. Chou, “Compression of 3D point clouds using a region-adaptive hierarchical transform,” \emph{IEEE Transactions on Image Processing}, vol. 25, no. 8, pp. 3947–3956, 2016.

\bibitem{ref51}
C. R. Qi, H. Su, K. Mo, and L. J. Guibas, “Pointnet: Deep learning on point sets for 3d classification and segmentation,” in \emph{Proceedings of the IEEE conference on computer vision and pattern recognition}, 2017, pp. 652–660.

\bibitem{ref52}
C. R. Qi, L. Yi, H. Su, and L. J. Guibas, “Pointnet++: Deep hierarchical feature learning on point sets in a metric space,” \emph{Adv Neural Inf Process Syst}, vol. 30, 2017.

\bibitem{ref53}
T.-Y. Lin, P. Goyal, R. Girshick, K. He, and P. Dollár, “Focal loss for dense object detection,” in \emph{Proceedings of the IEEE international conference on computer vision}, 2017, pp. 2980–2988.

\bibitem{ref54}
Z. Guo, Y. Zhang, L. Zhu, H. Wang, and G. Jiang, “TSC-PCAC: Voxel Transformer and Sparse Convolution-Based Point Cloud Attribute Compression for 3D Broadcasting,” \emph{IEEE Transactions on Broadcasting}, vol. 71, no. 1, pp. 154–166, 2025, doi: 10.1109/TBC.2024.3464417.

\bibitem{ref55}
X. Mao, H. Yuan, T. Guo, S. Jiang, R. Hamzaoui, and S. Kwong, “SPAC: Sampling-based Progressive Attribute Compression for Dense Point Clouds,” arXiv preprint arXiv:2409.10293, 2024.

\bibitem{ref56}
Y. Qian, J. Hou, S. Kwong, and Y. He, ``PUGeo-Net: A Geometry-Centric Network for 3D Point Cloud Upsampling,'' in \emph{Computer Vision -- ECCV 2020: 16th European Conference, Glasgow, UK, August 23--28, 2020, Proceedings, Part XIX}, 2020, pp. 752--769.

\bibitem{ref57}
R. Li, X. Li, C.-W. Fu, D. Cohen-Or, and P.-A. Heng, “Pu-gan: a point cloud upsampling adversarial network,” in \emph{Proceedings of the IEEE/CVF international conference on computer vision}, 2019, pp. 7203–7212.

\bibitem{ref58}
X. Fan, G. Li, D. Li, Y. Ren, W. Gao, and T. H. Li, “Deep geometry post-processing for decompressed point clouds,” in \emph{2022 IEEE International Conference on Multimedia and Expo (ICME)}, IEEE, 2022, pp. 1–6.

\bibitem{ref59}
L. Wang, J. Sun, H. Yuan, R. Hamzaoui, and X. Wang, “Kalman filter-based prediction refinement and quality enhancement for geometry-based point cloud compression,” in \emph{2021 International Conference on Visual Communications and Image Processing (VCIP)}, IEEE, 2021, pp. 1–5.

\bibitem{ref60}
J. Xing, H. Yuan, C. Chen, and T. Guo, “Wiener filter-based point cloud adaptive denoising for video-based point cloud compression,” in \emph{Proceedings of the 1st International Workshop on Advances in Point Cloud Compression, Processing and Analysis}, 2022, pp. 21–25.

\bibitem{ref61}
T. Guo, H. Yuan, R. Hamzaoui, X. Wang, and L. Wang, “Dependence-based coarse-to-fine approach for reducing distortion accumulation in G-PCC attribute compression,” \emph{IEEE Trans Industr Inform}, 2024.

\bibitem{ref62}
Y. Wei, Z. Wang, T. Guo, H. Liu, L. Shen, and H. Yuan, “High Efficiency Wiener Filter-based Point Cloud Quality Enhancement for MPEG G-PCC,” \emph{IEEE Transactions on Circuits and Systems for Video Technology}, p. 1, 2025, doi: 10.1109/TCSVT.2025.3552049.

\bibitem{ref63}
X. Sheng, L. Li, D. Liu, and Z. Xiong, “Attribute artifacts removal for geometry-based point cloud compression,” \emph{IEEE Transactions on Image Processing}, vol. 31, pp. 3399–3413, 2022.

\bibitem{ref64}
W. Liu, W. Gao, and X. Mu, “Fast inter-frame motion prediction for compressed dynamic point cloud attribute enhancement,” in \emph{Proceedings of the AAAI Conference on Artificial Intelligence}, 2024, pp. 3720–3728.

\bibitem{ref65}
D. Ding, J. Zhang, J. Wang, and Z. Ma, “Carnet: compression artifact reduction for point cloud attribute,” arXiv preprint arXiv:2209.08276, 2022.

\bibitem{ref66}
J. Zhang, J. Zhang, D. Ding, and Z. Ma, “Learning to restore compressed point cloud attribute: A fully data-driven approach and a rules-unrolling-based optimization,” \emph{IEEE Trans Vis Comput Graph}, vol. 31, no. 4, pp. 1985–1998, 2024.

\bibitem{ref67}
L. Gao, Z. Li, L. Hou, Y. Xu, and J. Sun, “Occupancy-assisted attribute artifact reduction for video-based point cloud compression,” \emph{IEEE Transactions on Broadcasting}, vol. 70, no. 2, pp. 667–680, 2024.

\bibitem{ref68}
A. Maggiordomo, F. Ponchio, P. Cignoni, and M. Tarini, “Real-world textured things: A repository of textured models generated with modern photo-reconstruction tools,” \emph{Comput Aided Geom Des}, vol. 83, p. 101943, 2020.

\bibitem{ref69}
C. Loop, Q. Cai, S. O. Escolano, and P. A. Chou, “Microsoft voxelized upper bodies - a voxelized point cloud dataset,” \emph{ISO/IEC JTC1/SC29 Joint WG11/WG1 (MPEG/JPEG)}, Geneva, Input document m38673/M72012, May 2016.

\bibitem{ref70}
E. d'Eon, B. Harrison, T. Myers, and P. A. Chou, “8i voxelized full bodies, version 2 – a voxelized point cloud dataset,” \emph{ISO/IEC JTC1/SC29 Joint WG11/WG1 (MPEG/JPEG)}, Geneva, Input document m40059/M74006, Jan. 2017.

\bibitem{ref71}
Y. Xu, Y. Lu, and Z. Wen, “Owlii dynamic human mesh sequence dataset,” \emph{ISO/IEC JTC1/SC29/WG11 MPEG}, Macau, Input document m41658, Oct. 2017.

\bibitem{ref72}
ISO/IEC, ``On balancing attribute QPs for GeSTM,'' ISO/IEC JTC1/SC29/WG7 MPEG M65830, Nov. 2023.

\bibitem{ref73}
G. Meynet, Y. Nehmé, J. Digne, and G. Lavoué, ``PCQM: A Full-Reference Quality Metric for Colored 3D Point Clouds,'' in \emph{2020 Twelfth International Conference on Quality of Multimedia Experience (QoMEX)}, 2020, pp. 1--6.

\bibitem{ref74}
Q. Liu, H. Su, Z. Duanmu, W. Liu, and Z. Wang, ``Perceptual quality assessment of colored 3D point clouds,'' \emph{IEEE Transactions on Visualization and Computer Graphics}, vol.~29, no.~8, pp.~3642--3655, 2022.

\bibitem{ref75}
C. Herglotz et al., ``The Bjontegaard Bible Why Your Way of Comparing Video Codecs May Be Wrong,'' \emph{IEEE Transactions on Image Processing}, vol.~33, pp.~987--1001, 2024, doi: 10.1109/TIP.2023.3346695.
\end{thebibliography}
\end{document}


\subsection{Subjective Quality Comparison for in Owlii Dataset}
\begin{figure*}[t]
    \centering
    \includegraphics[width=\linewidth]{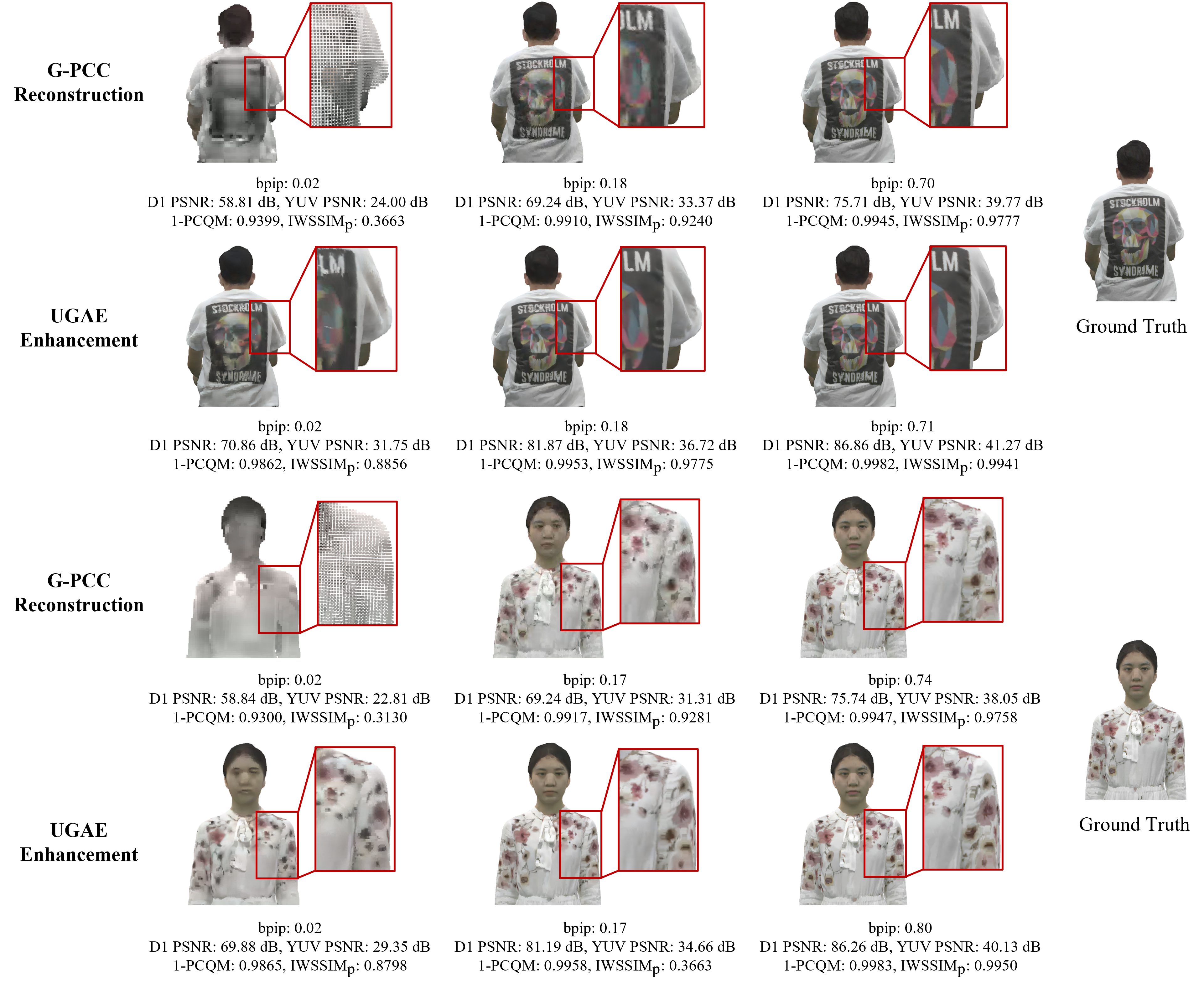}
    \caption{Subjective quality comparison. We selected three bitrates, R01, R03, and R05, to compare the reconstructed point clouds of G-PCC and UGAE.}
    \label{fig:visual_comparison2}
\end{figure*}

We conducted a comparative analysis of reconstructed point clouds from the Owlii dataset, focusing on both geometry and attributes, using G-PCC and UGAE. As illustrated in ~\ref{fig:visual_comparison2}, G-PCC reconstructions exhibit rough geometry and blurred attributes, whereas UGAE enhancements significantly enhances point clouds with smoother geometry and prominent high-frequency attribute details.

\label{fig:appendix_fig}

